\pdfoutput=1

\documentclass[11pt]{article}

\usepackage{acl}
\usepackage{float}
\usepackage{times}
\usepackage{latexsym}
\usepackage{booktabs}
\usepackage[T1]{fontenc}
\usepackage{booktabs}
\usepackage{multirow}
\usepackage[most]{tcolorbox}

\usepackage[utf8]{inputenc}
\usepackage{array}
\usepackage{microtype}

\usepackage{inconsolata}

\usepackage{graphicx}

\usepackage{amsmath}

\title{{\sc BeliefSim}: Towards Belief-Driven Simulation of\\ Demographic Misinformation Susceptibility}

\author{Angana Borah$^1$\hspace{5pt} 
 Zohaib Khan$^1$\hspace{5pt}  
 Rada Mihalcea$^1$\hspace{5pt}  
 Verónica Pérez-Rosas$^2$ \\
$^1$University of Michigan - Ann Arbor, USA  \\
$^2$ Texas State University  \\
\textit{\{anganab, 
zohaibkh, mihalcea\}@umich.edu} \hspace{5pt} \textit{vperezr@txstate.edu} \\  }

\begin{document}
\maketitle
\begin{abstract}
Misinformation is a growing societal threat, and susceptibility to misinformative claims varies across demographic groups due to differences in underlying beliefs. As Large Language Models (LLMs) are increasingly used to simulate human behaviors, we investigate whether they can simulate demographic misinformation susceptibility, treating \textit{beliefs} as a primary driving factor. We introduce {\sc BeliefSim}, a simulation framework that constructs demographic belief profiles using psychology-informed misinformation taxonomies and survey priors. We study prompt-based conditioning and post-training adaptation, and conduct a multi-fold evaluation using: (i) susceptibility alignment and (ii) counterfactual demographic sensitivity. Across both datasets and modeling strategies, we show that beliefs provide a strong prior for simulating misinformation susceptibility, with alignment up to 92\%. 


\end{abstract}

\section{Introduction}

Misinformation is a critical challenge in today’s information ecosystem, with impacts that vary substantially across demographic groups~\cite{khachaturov2025governments, timm2025tailored}. These differences reflect not only differential targeting~\cite{10.1145/3287560.3287580}, but also variation in how content is perceived~\cite{doi:10.1126/sciadv.aau4586}. Prior work illustrates these complexities: younger adults were more susceptible to believing COVID-related misinformation in the UK and Brazil~\cite{vijaykumar2021shades}, while older adults were more likely to share false articles on social media~\cite{guess2019less, brashier2020aging}. Conversely, prior studies also show that beliefs play a strong role~\cite{sultan2024susceptibility}: people are more likely to accept false information when it aligns or coincides with beliefs they already hold, a phenomenon known as ``belief consistency''~\cite{flynn2017nature, taber2006motivated, roozenbeek2020susceptibility}. Therefore, understanding how demographic differences relate to underlying beliefs is essential to explain  \textit{misinformation susceptibility}, 
i.e., how likely someone is to believe a misinformative claim.

\begin{figure}[t]
\centering
\includegraphics[width=\linewidth]{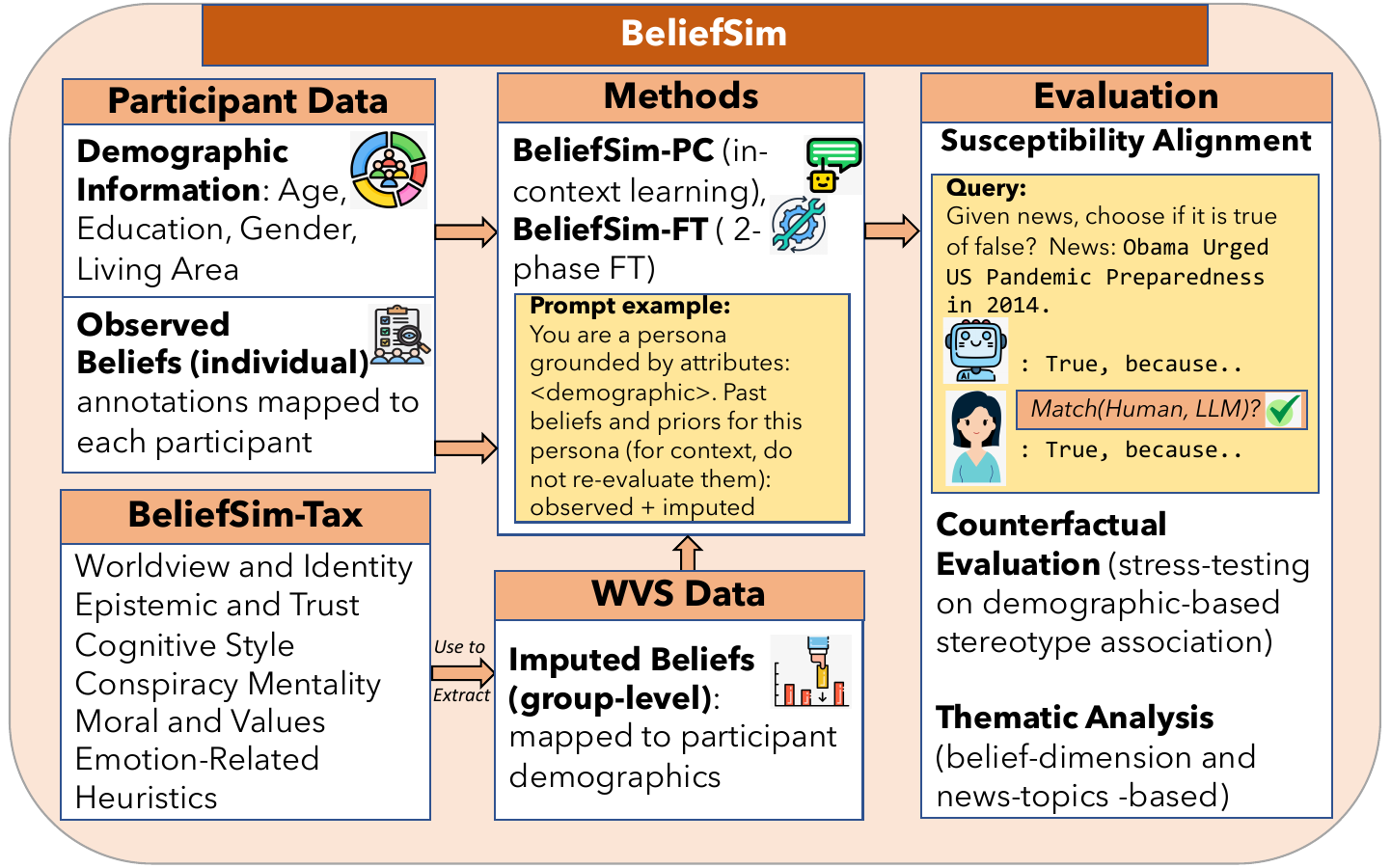}
\caption{\textbf{BeliefSim Framework.} (1) Participant Data, Observed and Imputed Beliefs (mapped to BeliefSim-Tax) are collected from surveys, (2) Methods consist of prompt-conditioning and post-training adaptation and (3) Evaluation is performed using Susceptibility Alignment, Counterfactual and Thematic Analysis.}
\label{fig:beliefsim}
\end{figure}

LLMs are increasingly used to simulate social processes~\cite{park2023generative, zhou2023sotopia, borah-etal-2025-mind}, including behaviors such as biased endorsement and echo chambers~\cite{acerbi2023large, nehring-etal-2024-large, borah2025persuasion, sharma2024generative}. Closest prior work on demographic misinformation susceptibility simulation relies on coarse personas~\cite{borah2025persuasion}, 
without modeling belief structures. 
Conversely, most belief-based studies primarily align LLMs with broad political ideologies or moral foundations~\cite{santurkar2023opinionslanguagemodelsreflect, Argyle_2023}, or simulate general personality traits~\cite{Pratelli_2025}. We believe that LLMs can serve as scalable testbeds for studying demographic susceptibility, but only if simulations move beyond surface-level personas toward belief-grounded representations.

\begin{table*}[t]
\centering
\scalebox{0.78}{
\begin{tabular}{p{0.10\linewidth} p{0.65\linewidth}p{0.43\linewidth}}
\toprule
\textbf{Dimension} & \textbf{Description} & \textbf{Example from WVS}\\
\midrule
Worldview \& Identity &
Group-based and political identities shape how individuals interpret and endorse information. & How proud are you to be from your country? \\
\midrule
Epistemic Trust &
Beliefs about what sources are credible and how knowledge should be evaluated (e.g., trust in science, media, institutions). & Would you say that the world is better off, or worse off, because of science? \\
\midrule
Cognitive Style &
Preferred thinking approach, ranging from analytical, accuracy-driven reasoning to intuitive impression-based judgments. &  Nowadays, does one often have trouble deciding which moral rules to follow? \\
\midrule
Conspiracy Mentality &
General tendency to perceive hidden plots or malevolent intent behind significant social or political events. & How many people who are in the media do you believe are involved in corruption? \\
\midrule
Morals \& Values &
Core moral values and cultural worldviews influence judgments of right, wrong, fairness, and societal norms. & Is it justifiable - claiming government benefits to which you are not entitled? \\
\midrule
Emotion-Related &
Emotion-driven beliefs related to fear, perceived risk, or threat sensitivity that heighten susceptibility to alarming misinformation. & How satisfied are you with your life as a whole these days? \\
\midrule
Heuristics &
Reliance on mental shortcuts (e.g., familiarity implies truth) that increase vulnerability to repeated or simplified claims. & Please indicate how much you use it: Daily newspaper, TV News, Radio News. \\
\bottomrule
\end{tabular}}
\caption{BeliefSim-Tax for modeling demographic-aware misinformation susceptibility in LLMs. We provide representative WVS items as examples of how survey questions align with each belief dimension.}
\label{tab:belief_taxonomy}
\end{table*}

Therefore, we ask three research questions: (1) Do beliefs improve demographic-aware simulations of misinformation susceptibility? (2) How can such simulations be rigorously evaluated using utility and counterfactual demographic analyses? and (3) What modeling strategies are  best suited for LLM-based simulation? 

To answer these questions, we propose \textsc{BeliefSim}, a belief-driven simulation framework for demographic misinformation susceptibility, illustrated in Fig~\ref{fig:beliefsim}. We summarize our contributions as follows: \textbf{(1)} We construct a belief taxonomy (\textsc{BelieSim-Tax}) and build a simulation dataset by aggregating data from existing surveys and studies; 
\textbf{(2)} We conduct empirical analyses to identify factors across belief and demographic dimensions that are most important for simulation; \textbf{(3)} We explore simulation techniques including prompt-based (\textsc{BeliefSim-PC}) and post-training (\textsc{BeliefSim-FT}) approaches using several LLMs; and \textbf{(4)} We perform a counterfactual study to understand when and how demographic information may provide useful priors vs stereotype-like sensitivity. Finally, we outline actionable steps that can help design future intervention methods.  


\section{Related Work}
Psychological research has established that susceptibility to misinformation is heavily driven by \textit{belief consistency}. While demographic factors like age, gender, and education provide significant predictive signals, recent large-scale meta-analyses suggest these effects can be context- and belief-dependent~\citet{sultan2024susceptibility}. However, most studies that establish benchmarks with large-scale human annotations face an inherent scalability bottleneck~\cite{maertens2024misinformation, borah2025persuasion}. High-quality human data collection is resource-intensive and slow, often limiting research to static snapshots (e.g. COVID-19) or narrow domains. 





The emerging capability of LLMs to simulate human behavior offers a potential solution. Early work showed that LLMs could act as ``silicon subjects'' via  demographic prompting~\cite{Argyle_2023}. Studies also investigated if LLMs reproduce classic social science findings and moral  values~\cite{park2023generative, borah-etal-2025-mind, nair2025language}. However, fidelity often degrades when simulations depend only on demographic labels~\cite{giorgi-etal-2024-modeling}, leading to stereotypical associations~\cite{borah-mihalcea-2024-towards}.



To move beyond demographics, recent research has focused on conditioning models with richer context such as beliefs~\cite{moon2024virtualpersonaslanguagemodels, namikoshi2024using}. Relevant relevant to our work, \citet{chuang2024demographicsaligningroleplayingllmbased} introduced ``Human Belief Networks'', showing that seeding an agent with a single belief (e.g., a stance on welfare) was more predictive of downstream responses than an entire demographic profile. However, most belief-focused studies have largely not examined their use in misinformation susceptibility. Our work bridges this gap by focusing on this domain, where the interplay between demographics and beliefs is highly complex.

\section{BeliefSim-Tax}
\label{sec:belief}
Beliefs are an important tool for modeling demographic simulations for misinformation. Prior work has shown that belief dimensions, such as conspiracy beliefs or trust in science, can help predict susceptibility to misinformation~\cite{ecker2022psychological, munusamy2024psychological}. 

Grounded in psychological and cognitive science research, we compiled \textsc{BeliefSim-Tax} (BeliefSim-Taxonomy) shown in Table ~\ref{tab:belief_taxonomy},  that includes seven core dimensions most associated with misinformation susceptibility: 
(1) Worldview and Identity Beliefs~\cite{kahan2017misconceptions, van2021speaking}, (2) Epistemic Trust Beliefs~\cite{de2021beliefs, lewandowsky2023misinformation}, (3) Cognitive style~\cite{,pennycook2019lazy, ecker2022psychological}, (4) Conspiracy mentality~\cite{douglas2019understanding, de2021beliefs}, (5) Moral and Value Beliefs~\cite{d2022personal, yang2024sharing}, (6) Emotion-Related Beliefs~\cite{brady2017emotion, mcloughlin2024misinformation} and (7) Heuristic Beliefs~\cite{lin2016social, fazio2020repetition}. While the above dimensions have traditionally been studied in isolation and within human populations, we unify them into a structured taxonomy designed for computational modeling (Further details are provided in Appendix~\ref{sec:taxonomy_belief}.
This enables systematic belief simulation by providing a framework for organizing belief data and evaluating it. 

\section{Data for Susceptibility Simulation}

\begin{figure}
\centering
\includegraphics[width=0.8\linewidth]{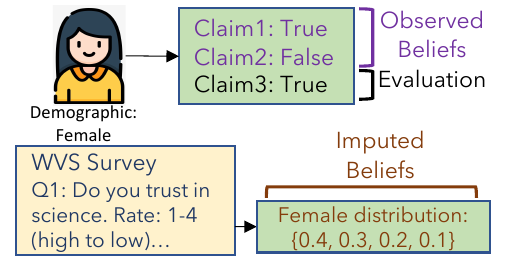}
\caption{\textbf{Data Example.} We map participant demographics to WVS responses to derive group-level imputed beliefs, and use claim-level evaluations as participant observed beliefs These are then used to predict observed beliefs.}
\label{fig:data_ex}
\end{figure}

For our study, we consider four demographic axes: Gender (female/male), Age (younger: <=35 years/older: >=60 years), Living Area (Rural/Urban), and Education (completed high school/not completed high school).\footnote{For ease of analysis, we focus on binary groupings, leaving finer-grained analysis to future work.} These demographics are commonly associated with systematic differences in media exposure, institutional trust, etc., all of which influence vulnerability to misinformation~\cite{allcott2017social, guess2019less, allcott2019trends, anspach2024not}. Focusing on these axes enables controlled analysis of demographic context for simulating susceptibility signals in LLMs.  


\subsection{Evaluation Data}

For susceptibility evaluation, we use two datasets containing human judgments of whether they believe a given claim: (1) PANDORA Dataset from~\citet{borah2025persuasion}, containing annotations from 318 participants. Each participant provided judgments on 3 distinct claims, along with demographic information: age, gender, living area, and education. The claims are collected from RumorEval~\cite{gorrell-etal-2019-semeval}, which consists of true/false rumors covering eight news and natural disaster events; (2) MIST dataset from~\citet{maertens2024misinformation}. From MIST, we use only Study 1 (MIST-1) which includes 409 participants, each providing judgment on the same 100 claims, and demographic information including age, gender, and education. 
From the two datasets combined, we obtain 13.8K claims for evaluation.


\subsection{Belief Data}
We collect two complementary belief signals, consisting of individually observed belief judgments and group-level (demographic) belief distributions:

\noindent \textbf{(1) Observed Data} -
claims that were directly judged by participants in the PANDORA and MIST-1 datasets. These responses capture individual-level belief judgments, reflecting each participant's personal stance rather than demographic group averages. For each evaluation instance, we include two held-out claim judgments from the same participant as observed beliefs, keeping them separate from the target evaluation claim. Across 13.8K evaluation instances, this yields 27.6K observed belief-judgment instances.

\noindent \textbf{(2) Imputed Data} - inferred from the World Values Survey Wave 7\footnote{\url{https://www.worldvaluessurvey.org/WVSDocumentationWV7.jsp}} distributions. Imputed data represent demographic belief priors (group–level), inferred from WVS, conditioned solely on demographic attributes. We map these imputed belief items to our belief taxonomy using exploratory factor analysis. This yields 126 imputed belief questions and corresponding demographic distributions. Table~\ref{tab:belief_taxonomy} shows examples, and Appendix~\ref{sec:wvs} contains mapping details and all questions.

\noindent Fig.~\ref{fig:data_ex} provides an example of simulation data. We restrict all our analysis to U.S.-based participants and English headlines, consistent with the scope of the available datasets, while still leveraging variation in living area, age, gender, and education.







\section{BeliefSim-PC: Prompt-based Conditioning}

We propose \textsc{BeliefSim-PC} (BeliefSim-Prompt-Cond.), which consists of prompts that condition the model on participant demographics and belief signals (both observed and imputed) from the surveys. Note that we simulate each demographic axis separately, as it enables controlled analysis of demographic-level differences.







\noindent \textbf{Method.} For WVS questions, answers are either Likert scales~\cite{bertram2007likert} or Yes/No questions. Therefore, we input modal (most frequent) responses in the prompt as beliefs. For example, consider the WVS question ``How important is religion in your life?'', rated on a 1–10 scale. If a demographic group most frequently responds with a value of 3, we use this modal response in the prompt. Given demographic group <d>, and belief annotations <b>, we use: \texttt{You are a persona grounded by attributes: <d>.  Past beliefs and priors for this persona (for context, do not re-evaluate them): <b>. When judging a claim, stay consistent with this persona's prior beliefs where reasonable.} We focus on modal responses in the main experiments for prompt stability, and include a robustness check with full belief-distribution prompting in Appendix~\ref{sec:dis_prompt}.



We perform experiments under four primary input conditions for prompt-based conditioning: (1) zero-shot (without any demographic or belief information), (2) demographic information only, (3) belief information only, and (4) both demographics and beliefs. We additionally ablate over belief dimensions (the seven in our taxonomy) and belief sources (observed vs. imputed), as well as over demographic attributes (analyzing groups separately). These experiments yield several insights across 12 total settings.  We perform these experiments using several instruction-tuned LLMs: \texttt{Llama-3-8B-Instr}, \texttt{Llama-3.2-3B-Instr}, \texttt{Qwen2.5-14B-Instr}, \texttt{Mistral-7B-Instr-v0.2}, \texttt{OLMo-2-7B-Instr}, \texttt{DeepSeek-LLM-7B-Chat}, and \texttt{Grok-4-Fast}. We report averaged results across three separate runs.

\noindent \textbf{Evaluation.} We measure susceptibility alignment: whether the LLM correctly predicts if an individual would judge a claim as true or false. We also discuss demographic divergence and the role of human judgment variability in Appendices~\ref{app:distribution_alignment} and~\ref{app:human_variability}.


\subsection{Results}
\begin{figure}
\centering
\includegraphics[width=0.8\linewidth]{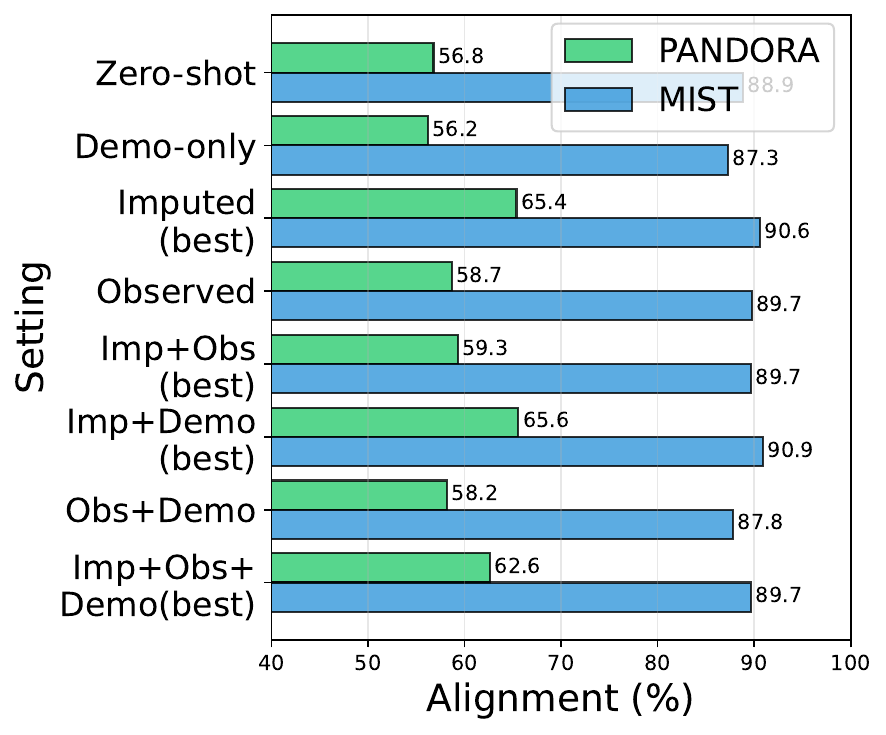}
\caption{\small \textbf{Susceptibility Alignment}, averaged across demographic groups and models. Imputed + Demo(graphic) (best) performs the best. All belief-based settings do better than zero-shot and demo-only. (Fig~\ref{fig:overall_comp_all} shows results of all settings.)}
\label{fig:overall_comp}
\end{figure}

Fig~\ref{fig:overall_comp} shows the susceptibility alignment across the belief and demographic configurations for both datasets. For cases that include \texttt{Imputed} beliefs, we only report \texttt{best} results by selecting the belief dimension that achieves the highest accuracy. Findings show that incorporating belief information consistently improves performance over the zero-shot and demographic-only baselines. We next analyze the impact of specific belief types, demographic ablations, datasets, and model performances. Appendix~\ref{sec:promptbased} contains detailed results for all settings. Furthermore, Appendix~\ref{sec:dis_prompt} shows that full belief-distribution prompts do not provide stable results across models. So, we use modal responses as a more stable and compact representation in the main combined settings.

\noindent \textbf{Demographic vs Belief information.} Overall, beliefs drive most of the positive gains. Demo-only degrades zero-shot performances for both datasets, while beliefs-only (imputed) show larger gains (~+1-16) points over demo-only and zero-shot settings. Imputed + Demo achieves the highest performance. This shows that what people \textit{believe} may be more predictive than \textit{who they are}. This is also in line with findings from~\citet{borah-etal-2025-mind}, which show that belief-based signals are stronger than demographic signals in LLMs.

\noindent \textbf{Imputed vs Observed Beliefs.}
Across settings, imputed beliefs consistently outperform observed beliefs, more so for PANDORA. A likely reason is that observed beliefs, being tied to specific participants, can be sparse, or mismatched to the new claim, while imputed demographic beliefs may act as \textit{smoother population priors}. Furthermore, combining imputed and observed beliefs helps, but remains below imputed-only, likely because sparse individual signals can conflict with or dilute the smoother demographic prior. Thus, accuracy trends are: $imputed>imputed+observed>observed$.

\noindent \textbf{Imputed Belief Ablations.} Adding one belief dimension at a time to the prompt consistently yields higher performance than including all belief dimensions together. Across datasets, emotion-related and moral-values are the strongest. These also vary by demographics: emotion-related is the strongest in Gender; moral-values is the strongest for Rural/Urban and Education; heuristics/cognitive beliefs are the strongest in Education (Detailed results are in Appendix~\ref{sec:belief_ana}). 

\noindent \textbf{Demographic Ablations.} Adding demographic information on top of beliefs yields small gains (in the range ~0.2-1.2\%), especially for imputed beliefs. Thus suggesting that demographics serve as an additional context signal, helping with relevant belief priors. 
On the effect of belief across demographic groups, accuracies for age, education and living area are improved with belief addition, specifically using the imputed beliefs. Gender shows the largest sensitivity to which beliefs are included; prior work also show that gender effects are often smaller/mixed in misinformation contexts~\cite{sultan2024susceptibility}. This underscores the need for caution in selecting belief evidence for gender-specific cases. 



\definecolor{CustomPastelPink}{RGB}{250, 180, 180}
\definecolor{CustomGreen}{RGB}{0, 180, 150}
\definecolor{CustomBlue}{RGB}{173, 216, 230}
\definecolor{CustomPastelYellow}{RGB}{253, 253, 150}


\begin{table}[t]
  \centering
  \scalebox{0.8}{
  \setlength{\tabcolsep}{4pt}
  \begin{tabular}{lcc}
  \hline
  \textsc{Model} & \textsc{PANDORA} & \textsc{MIST} \\
  \hline
  Qwen2.5-14B & \colorbox{CustomBlue}{62.28} & \colorbox{CustomBlue}{95.97} \\
  Llama-3-8B  & 57.49 & 90.53 \\
  Llama-3.2-3B   & 53.89 & 85.14 \\
Grok-4-Fast    & 57.07 & 87.20 \\
  Mistral-7B-v0.2     & 55.99 & 70.07 \\
  OLMo-2-7B           & 50.60 & 76.95 \\
  DeepSeek-7B         & 52.40 & 59.04 \\
  \hline
  \end{tabular}
  }
  \caption{\textbf{Model performance.} Qwen performs the best followed by Llama models while Deepseek has the worst alignment.}
  \label{tab:model_comparison}
  \end{table}

\noindent \textbf{Model and Dataset comparison.} Across models, Qwen consistently shows higher accuracies, followed by other models, as shown in Table~\ref{tab:model_comparison}. 

Across datasets, MIST yields a high zero-shot baseline of 88.9\% and gains the most from belief integration in the best imputed + demographic setting, reaching 90.9\% (+2.1). In contrast, PANDORA has a lower zero-shot baseline of 56.8\%, but shows a larger gain from belief integration, reaching 65.6\% in the best imputed + demographic setting (+8.8). Therefore, the impact of adding belief priors varies across datasets: PANDORA contains fewer examples than MIST (318 vs. 13.5K), and benefits more from belief conditioning. Nevertheless, adding belief priors is beneficial across both datasets after screening outlier modal runs.

We further compare veracity/factuality accuracy with susceptibility simulation in Appendix~\ref{app:veracity_vs_susceptibility}, showing that factual correctness and participant-judgment prediction are distinct objectives: models with stronger veracity accuracy do not always better predict human judgments. We also test whether the remaining gap is due to pretraining knowledge overriding persona conditioning (Appendix~\ref{sec:factual-confidence}). We find that factual confidence biases zero-shot predictions, but is not the dominant limiting factor.




\noindent \textbf{Thematic Analysis.} We cluster news claims into latent topics and test whether LLM susceptibility alignment differs across demographics by topic. We conduct our analysis on the MIST dataset, as it is much larger than PANDORA. We find that demographic differences in susceptibility are topic-dependent: clusters corresponding to lexically obvious or opinion-based claims exhibit minimal demographic variation, whereas more ambiguous or diverse topics like science/health claims and government show larger differences across age (older groups scoring higher) and education groups (higher-educated groups scoring higher). Gender-based differences are consistently small across all topics. These findings suggest that demographic susceptibility is also contextual and topic-sensitive, motivating further evaluation in future studies (More details are in Appendix~\ref{sec:thematic}). 



\subsection{Demographic-based Counterfactual Evaluation}
\label{sec:counterfactual-subsec}
Results show that demographics can be a helpful context signal in addition to belief information and legitimately correlate with misinformation susceptibility. But they can also induce spurious signals that appear accurate but could be stereotypical~\cite{wan2025truthtricksmeasuringmitigating, geirhos2020shortcut}. To investigate this, we conduct counterfactual evaluations to assess whether demographic attributes provide informative priors or introduce stereotype-sensitive associations.
\begin{figure}
\centering
\includegraphics[width=0.8\linewidth]{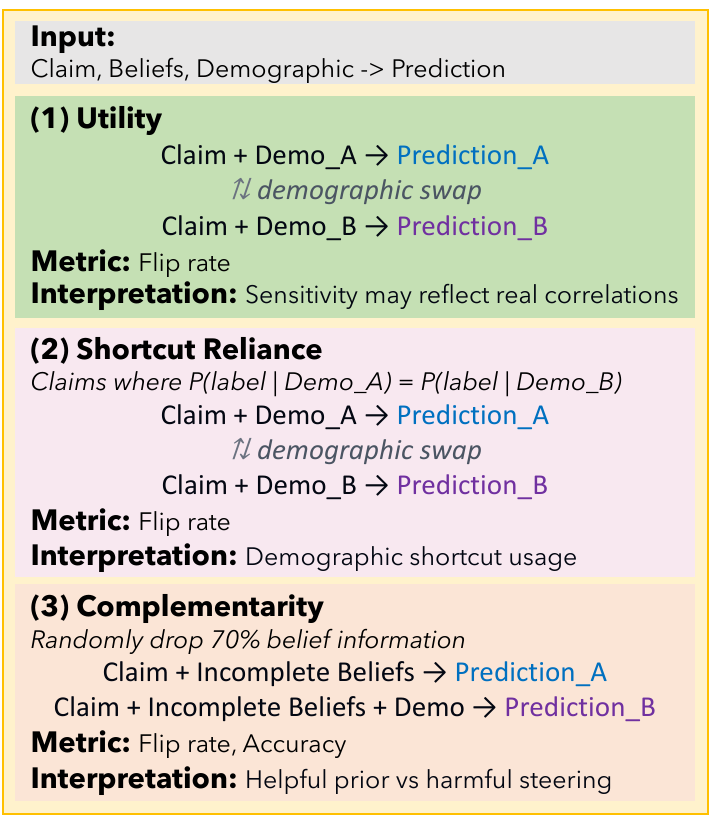}
\caption{\textbf{Demographic-based Counterfactual Evaluation.} Note that we perform a demographic swap within the same group, i.e., swap male with female, rural with urban, etc. keeping the claim constant.}
\label{fig:counter}
\end{figure}

\noindent \textbf{Method.} We conduct three complementary analyses (Fig~\ref{fig:counter}): 

\noindent \textit{(1) Utility}: Do demographics improve susceptibility prediction beyond claim content? We test utility first with demographic-only prompts, then swap the demographic attribute while keeping the claim fixed and measure prediction flips. Higher flip rates indicate stronger demographic sensitivity, which may reflect real demographic-label correlations rather than only spurious shortcuts.

\noindent \textit{(2) Shortcut reliance}: Does the model rely on demographic cues even when they are non-informative by construction? We create a controlled subset where each claim has matched human-label distributions across demographic groups, making demographics non-predictive. High flip rates here indicate short-cutting and possible stereotyping.


\noindent \textit{(3) Complementarity}: When belief evidence is incomplete, do demographics help or hurt? We simulate missing evidence by dropping 70\% of total WVS belief statements, then compare belief-only vs. belief+demographic predictions. Flip rates show how much demographics steer predictions under uncertainty, while accuracy indicates whether this steering improves performance.


\begin{figure}
\centering
\includegraphics[width=0.9\linewidth]{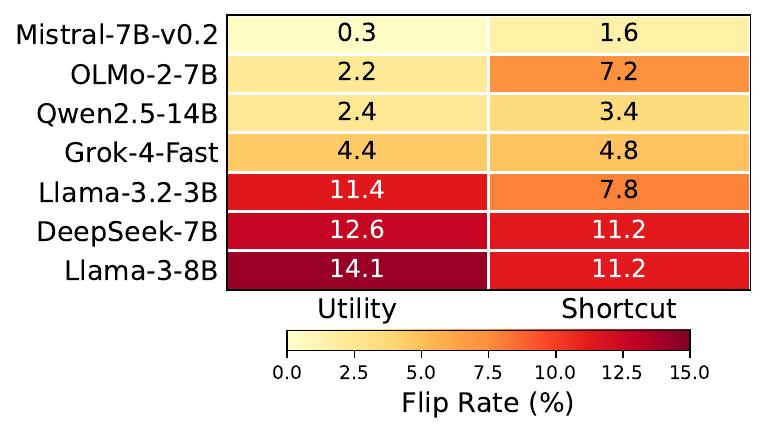}
\caption{\small \textbf{Flip-Rates for Counterfactual Evaluation}: Mistral and Qwn models have lower flip-rates whereas Llama and Deepseek have higher rates. Darker (red) colors mean higher flip rates while lighter (yellow) colors mean lower flips.}
\label{fig:flip_rates}
\end{figure}

\noindent \textbf{Results.} Fig~\ref{fig:flip_rates} shows the flip-rates per model averaged across demographic groups and datasets. Overall, \texttt{Mistral} shows the lowest flip rates, suggesting limited dependence on demographic cues. \texttt{Qwen} is also comparatively stable and low. In contrast, \texttt{Llama} and \texttt{Deepseek} models show substantially higher demographic sensitivity. Across models, the largest flip rates concentrate in Education and Living Area, especially in shortcut reliance, which could relate to stereotype-driven reliance (details and qualitative evidence in Appendix~\ref{app:flip_rates}).


\begin{figure}[ht]
\centering
\includegraphics[width=0.9\linewidth]{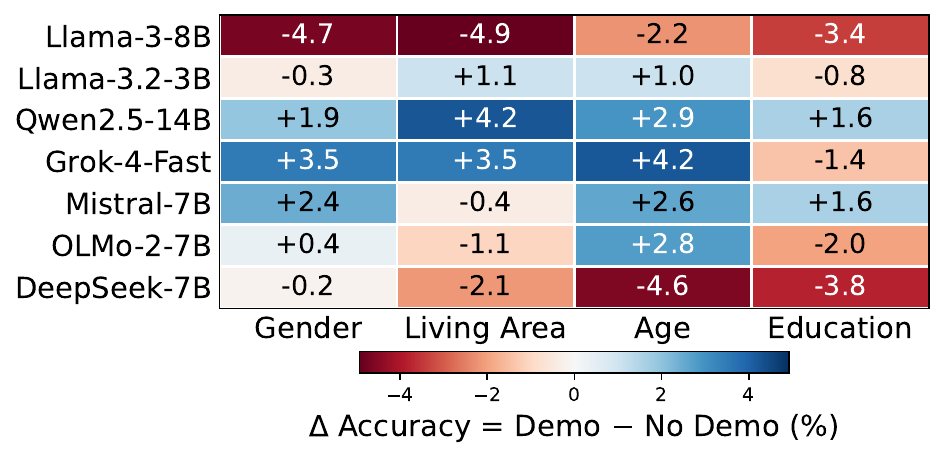}
\caption{\small \textbf{Complementarity Analysis.} Adding demographics has mixed effects (mostly negative) when partial belief information is added. Positive values mean that adding demographics increases accuracy, while negative values mean that it decreases accuracy.}
\label{fig:complement}
\end{figure}

For complementarity analysis, Fig~\ref{fig:complement} shows that demographics show a mixed effect. \texttt{Qwen} improves for all axes while \texttt{Mistral} and \texttt{Grok} also show mostly positive gains (except one axis). However, other models show mixed or negative shifts, especially for Living Area and Education. 

Overall, these findings suggest that demographic sensitivity is highly model-dependent and demographics provide limited practical benefit when partial beliefs are present, showing they are a weak prior at best. This motivates moving beyond prompt-only conditioning, as demographic cues in the prompt can introduce unstable shortcuts. Therefore, belief signals must be integrated in a way that is robust to conflicting sources.

\section{BeliefSim-FT: Post-training Adaptation}
Prompt-based conditioning experiments show the benefits of using beliefs for demographic misinformation susceptibility simulation. However, mixing imputed and observed beliefs in the prompt is associated with lower performance. This is intuitive, as observed (ground-truth) beliefs are tied to individual participants and imputed beliefs are only linked to demographic groups and not individuals, creating inconsistencies. Furthermore, training directly on demographics, modal responses and susceptibility labels risks leakage, where the model may learn label shortcuts as observed in counterfactual evaluation~\cite{wan2025truthtricksmeasuringmitigating,geirhos2020shortcut}. To address this, we propose \textsc{BeliefSim-FT} (BeliefSim-Fine-Tuning), which decouples imputed beliefs from observed data, while still learning transferable belief representations through a separate adapter.


\noindent \textbf{Method.} To decouple group-based imputed beliefs from individualistic observed beliefs, BeliefSim-FT consists of a two-phase design: (1) belief modeling, and (2) susceptibility fine-tuning. 

\textbf{Phase 1: Belief Modeling.} We train a belief adapter by freezing the base LLM and training a belief head, implemented as a linear projection followed by softmax, to predict demographic-conditioned belief distributions for WVS questions. Concretely, given the encoder representation $h_{\phi}(q,d)$ for question $q$ and demographic context $d$, we model $p_{\theta}(y\mid q,d)=\mathrm{softmax}(Wh_{\phi}(q,d)+b)$. This captures population-level variability in belief responses. Importantly, this stage uses only survey supervision (no misinformation labels), reducing the risk of label leakage or reward hacking. Because BeliefSim-FT separates belief modeling from susceptibility prediction, extending to additional personas primarily requires collecting additional demographic-belief distributions, while the same adapter-based architecture can be reused.


\textbf{Phase 2: Susceptibility Fine-tuning.}
We then freeze the belief adapter and train a lightweight susceptibility head that combines the base LLM’s semantic representation of the prompt with the belief representation produced by the adapter. Given a persona-conditioned input (demographics and observed beliefs) and target claim $x$, the frozen adapter outputs a belief embedding $z_{\text{bel}}$, which we concatenate with the claim representation $h_{\phi}(x)$ and predict susceptibility via a binary classifier:
\begin{equation}
\small
p_{\psi}(y\mid x,d)=\mathrm{softmax}\!\left(U\,[\,h_{\phi}(x)\,;\,z_{\text{bel}}(d)\,]+c\right),
\end{equation}
where $y\in\{\text{true},\text{misinformation}\}$. We train only the susceptibility head parameters $\psi=\{U,c\}$ with cross-entropy loss, keeping the base model and belief adapter frozen. This is to ensure susceptibility learning relies on transferable belief structure rather than encoding demographic shortcuts.

\begin{figure}
\centering
\includegraphics[width=0.9\linewidth]{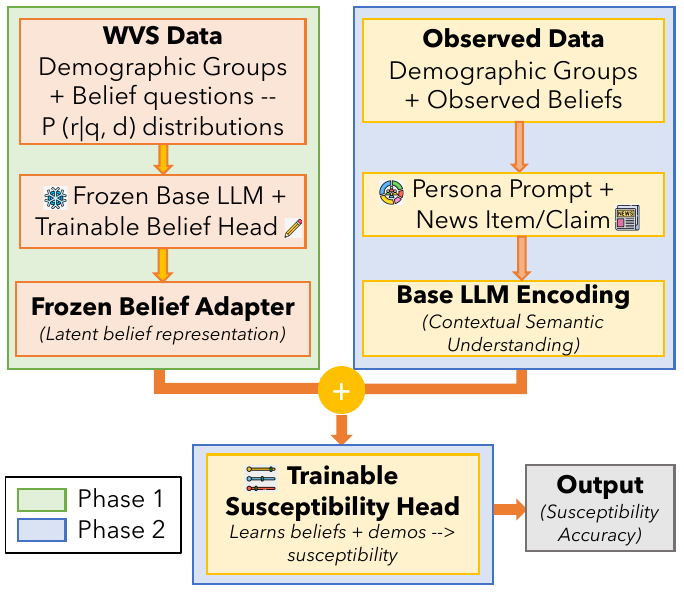}
\caption{\small \textbf{BeliefSim-FT framework.} Green-shaded components correspond to Phase 1 (Belief Modeling), and blue-shaded components are Phase 2 (Susceptibility Fine-Tuning).}
\label{fig:BeliefSim-FT}
\end{figure}
\noindent \textbf{Data.} For Phase 1, we construct demographic belief priors from empirical WVS response distributions. 
For each question $q$ and demographic group $d$, we estimate a categorical distribution $P(r \mid q, d)$ over response options $r \in \{1,\dots,K_q\}$. 
These distributions capture meaningful demographic variability across WVS belief questions. We restrict our experiments to six demographic groups, as MIST does not include living-area annotations, and the PANDORA data is too small to support standalone data for fine-tuning. This results in $126$ WVS response distributions per demographic group, yielding $6 \times 125$ $= 750$ examples in total. We apply textual augmentation via back-translation using Google Translate,\footnote{https://translate.google.com/} producing $1250$ examples. This is sufficient as the belief adapter is a lightweight linear head requiring fewer examples to reliably capture demographic belief patterns. Training is performed using an 80/20 train/val split. 

For Phase 2, we use (PANDORA+MIST-1) data, keeping each demographic group separate per data point (as in prompt conditioning). The target is the prediction on a claim, giving susceptibility scores. We use a 80/20 train/val split, yielding approximately ~33.1k training examples and ~5.1k validation examples (after removal of overlapping examples). We also use MIST-2b as a cross-study, cross-participant evaluation under the standardized MIST instrument consisting of 7k evaluation points. Appendix~\ref{app:mist_overlap} discusses the distinction between participant leakage and claim overlap.






\definecolor{CustomPastelPink}{RGB}{250, 180, 180}
\definecolor{CustomGreen}{RGB}{0, 180, 150}
\definecolor{CustomBlue}{RGB}{0, 100, 200}
\definecolor{CustomPastelYellow}{RGB}{253, 253, 150}

\begin{table}[h]
\tiny
\centering
\resizebox{\columnwidth}{!}{%
\begin{tabular}{lccccc}
\toprule
\multirow{2}{*}{\textbf{Model}} & \multicolumn{2}{c}{\textbf{PANDORA+MIST-1}} & \multicolumn{2}{c}{\textbf{MIST-2}} \\
\cmidrule(lr){2-3} \cmidrule(lr){4-5}
& \textbf{Acc} & \textbf{F1} & \textbf{ Acc} & \textbf{F1} \\
\midrule
\multicolumn{5}{c}{\textbf{Baseline (1-phase)}} \\
\midrule
Llama-3-8b & 0.750 & 0.751 & 0.680 & 0.673 \\
Qwen-2.5-14b & 0.760 & 0.761 & 0.720 & 0.712 \\
Mistral-7B & 0.740 & 0.761 & 0.650 & 0.653 \\
\midrule
\multicolumn{5}{c}{\textbf{LoRA-FT (2-phase)}} \\
\midrule
Llama-3-8b & \colorbox{CustomPastelPink}{0.809} & \colorbox{CustomPastelPink}{0.809} & 0.876 & 0.861 \\
Qwen-2.5-14b  & 0.800 & 0.800 & 0.893 & 0.892 \\
Mistral-7B & 0.800 & 0.790 & 0.887 & 0.867 \\
\midrule
\multicolumn{5}{c}{\textbf{BeliefSim-FT (2-phase)}} \\
\midrule
Llama-3-8b & 0.793 & 0.793 & 0.884 & 0.896 \\
Qwen-2.5-14b & 0.792 & 0.791 & \colorbox{CustomPastelPink}{0.924} & \colorbox{CustomPastelPink}{0.906} \\
Mistral-7B & 0.792 & 0.792 & 0.884 & 0.823 \\

\bottomrule

\end{tabular}%
}
\caption{\textbf{Fine-tuning performances. }2-phase belief training improves performance across both datasets.}
\label{tab:ft}
\end{table}

We compare BeliefSim-FT against two baselines: standard one-phase LoRA fine-tuning on PANDORA+MIST-1, and a two-phase LoRA variant that replaces our head-based adaptation. We evaluate three representative models: \texttt{Qwen} (high alignment, low shortcut reliance), \texttt{Mistral} (low alignment, low shortcut reliance), and \texttt{Llama} (high alignment, high shortcut reliance).

\noindent \textbf{Evaluation Metrics.} Phase 1 reflects population-level belief fidelity, so we use distribution-level metric over Likert-scale responses, Kullback--Leibler (KL) divergence~\cite{kullback1951information}. This measures how well predicted belief distributions align with each demographic (lower is better for both). Phase 2 reflects task utility. So, we compute susceptibility alignment for this phase.



\subsection{Results} 

\noindent \textbf{Phase 1.} \texttt{Qwen} has the lowest KL-divergence (0.051), followed by \texttt{Mistral} (0.087) and \texttt{Llama} (0.287). With LoRA-ft, the trends remain similar (complete results are in Appendix~\ref{sec:ft_ana}).

\noindent \textbf{Phase 2.} Table~\ref{tab:ft} shows that cross-study transfer to MIST-2 improves substantially when moving from 1-phase to 2-phase training. The 1-phase baseline performs well on PANDORA+MIST-1 but drops on MIST-2, while BeliefSim-FT achieves the strongest MIST-2 performance (up to 92.4\%). The higher MIST-2 scores may partly reflect that it uses the standardized MIST instrument with reliability-selected items, whereas PANDORA+MIST-1 contains broader, more heterogeneous claims. LoRA-FT slightly improves on PANDORA+MIST-1 but transfers less strongly, possibly due to mild overfitting from updating more parameters. Across models, \texttt{Qwen} performs best in the 2-phase settings. Shortcut-reliance experiments further show that BeliefSim-FT reduces flip rates to nearly 0\%, even for \texttt{Llama}, which had the highest flip rate in the prompt-based setting (Appendix~\ref{sec:shortcut}).








\section{Lessons Learned and Actionable Steps}
Our findings show that belief priors are central to simulating demographic misinformation susceptibility with LLMs. We introduce BeliefSim, and show that decoupling belief modeling from susceptibility prediction enables higher accuracy while reducing spurious demographic sensitivity. These findings offer actionable insights to design targeted demographic interventions.

\noindent \textbf{Beliefs are crucial to demographic susceptibility simulation.} Adding belief priors (especially imputed) leads to high accuracy gains across. Future studies can investigate additional survey sources to broaden coverage and focus on diverse demographic contexts. Furthermore, studying the effective decoupling of belief types is important. 

\noindent \textbf{Simple head tuning is effective for simulation alignment in smaller models.} Our two-phase BeliefSim-FT approach generalizes better than full fine-tuning, suggesting that lightweight, modular adaptation can be effective while being cheaper. Future work can improve robustness by exploring alternative adapter designs and fusion mechanisms for combining belief and text representations.

\noindent \textbf{Counterfactual evaluation is important for simulation studies.} Counterfactual experiments quantify each model's reliance on demographic shortcuts. Such measures are necessary as demographic correlation can be a real/spurious cue. Future work can build on this by developing fine-grained stress tests to understand what drives model flips.

\section{Conclusion}
This paper introduced BeliefSim, a belief-driven framework for simulating demographic misinformation susceptibility using LLMs. Across two datasets and through prompt-based conditioning and post-training adaptation, we show that belief priors are a key driver for demographic-level susceptibility patterns. In contrast, demographics alone are an unreliable signal that can induce shortcut reliances. Our results highlight the importance of decoupling imputed and observed belief sources and we provide practical evaluation tools for it, such as prediction accuracy and counterfactual sensitivity. Based on these findings, we outline directions for future work and release our open-source framework, BeliefSim.\footnote{\url{https://anonymous.4open.science/r/belief-sim}}




\section*{Limitations}

\subsection*{Coverage and Intersectionality}

Our demographic modeling is intentionally largely single-axis: we evaluate on a small set of demographic attributes (8 groups) and treat them independently. Our goal was to isolate the effects of simulation across demographic groups and simplify counterfactual analysis as well as balance labels across groups. However, intersectional groups (e.g., older x low-education, female x rural) can have interesting implications for simulation purposes, and future work can further investigate BeliefSim for intersectional belief profiles. In addition, while we mostly adhere to binary demographic axes, each axis can be further analyzed in a finer-grained approach. Finally, our study is based on US participants only, due to the availability of current data sources. We acknowledge that these may have WEIRD implications~\cite{mihalcea2025ai}, and future work should focus on investigating wider, cross-cultural misinformation susceptibility.

\subsection*{Counterfactual Flips do not always mean stereotyping}
Our flip-rate analyses quantify sensitivity to demographic perturbations, but flips do not uniquely measure stereotype-driven shortcutting. Furthermore, low flip rates do not guarantee fairness - models may still encode demographic effects indirectly through correlations or beliefs. Although our experiments help isolate spurious dependence, our experiments may not capture real-world causal pathways as they require much extensive analysis. Future work can investigate extending our counterfactual flips experiments with stronger causal/robustness evaluations to better understand causal demographic effects and/or measure stereotype-driven shortcutting. 

\subsection*{Experimental Limitations}
Access to closed-source LLMs is constrained by transparency and cost. Therefore, we limit our experiments to three open-source models (to support controlled counterfactual swaps, ablations, and multiple runs). As a result, our conclusions may not fully generalize to the strongest closed-source models. Future work can extend our framework by benchmarking a broader set of closed-source models under matched prompting and budgeted sampling.

\section*{Acknowledgments}

We thank the anonymous reviewers for their constructive feedback. We are also grateful to the members of the Language and Information Technologies Lab at the University of Michigan for their valuable input and insightful discussions during the early stages of the project. This project was funded by award \#80345 from the Robert Wood Johnson Foundation. Any opinions, findings, and conclusions or recommendations expressed in this material are those of the authors and do not necessarily reflect the views of the Robert Wood Johnson Foundation.
\bibliography{custom}

\appendix


\section{Taxonomy Dimensions and Imputed Beliefs}
\label{sec:taxonomy_belief}

Beliefs are an important tool for modeling demographic simulations in the context of misinformation. Psychological work shows that several belief dimensions, such as conspiracy beliefs, political ideology, trust in science, etc. can help us predict who is most susceptible to misinformation~\cite{ecker2022psychological, munusamy2024psychological}. Rather than using only demographic information as a flat category, adding belief profiles provides a more fine-grained representation of how different groups may perceive information. Recent LLM studies also argue for modeling population-level beliefs and preferences to simulate targeted groups, using belief-like representations to approximate human responses at scale~\cite{namikoshi2024using, kaiser2025simulating}. 

\subsection{Taxonomy formation}
We create a belief taxonomy of beliefs consisting of seven core dimensions that are associated with misinformation susceptibility grounded in psychological and cognitive science research: 

\noindent \textbf{(1) Worldview and Identity Beliefs}: individuals interpret information through the lens of social identity and worldview~\cite{kahan2017misconceptions, van2021speaking}, \textbf{(2) Epistemic Trust Beliefs}: individuals differ systematically in epistemic trust toward institutions and experts~\cite{de2021beliefs, lewandowsky2023misinformation}, \textbf{(3) Cognitive style}: individuals vary in cognition, such as reliance on analytic versus intuitive reasoning~\cite{,pennycook2019lazy, ecker2022psychological}, \textbf{(4) Conspiracy mentality}: individuals differ in a generalized predisposition to conspiracies~\cite{douglas2019understanding, de2021beliefs}, \textbf{(5) Moral and Value Beliefs}: individuals prioritize different moral values~\cite{d2022personal, yang2024sharing}, \textbf{(6) Emotion-Related Beliefs}: individuals vary in emotional responsiveness (heightened emotional arousal, such as anger or fear may amplify belief in misinformation)~\cite{brady2017emotion, mcloughlin2024misinformation} and \textbf{(7) Heuristic Beliefs}: individuals rely to different degrees on shortcuts such as repetition, familiarity, and social endorsement~\cite{lin2016social, fazio2020repetition}. 

While the above dimensions have traditionally been studied in isolation and within human populations, we unify them into a structured taxonomy designed for computational modeling. This taxonomy enables systematic belief simulation in models by providing a framework for collecting and organizing belief data and providing interpretable axes along which belief priors can be instantiated in models. 


\subsection{Mapping to Belief Taxonomy.}
\label{sec:wvs}
We map imputed data to the above belief dimensions. We apply exploratory factor analysis on WVS responses to identify latent belief groupings, examining which items naturally cluster together (e.g., trust-related items to Epistemic \& Trust Beliefs). Post that, we conduct a manual review to ensure proper alignment with our taxonomy and validate the final question-to-dimension mapping. Overall, we obtain 126 imputed belief questions across dimensions. 
Table~\ref{tab:wvs_questions_mapped} contains the 126 WVS questions mapped to the 7 belief taxonomy dimensions. 


\section{Datasets}
\label{sec:dataset}
\subsection{Evaluation Data}

For susceptibility evaluation, we use two ground-truth datasets containing human judgments of whether they believe a given claim: (1) PANDORA Dataset from~\citet{borah2025persuasion}, containing annotations from 318 participants. Each participant provided judgments on 3 distinct claims, along with demographic information including age, gender, living area, and education. These claims are collected from RumorEval~\cite{gorrell-etal-2019-semeval}, which consists of true or false rumors covering eight major news events and natural disaster events; (2) MIST  dataset from~\citet{maertens2024misinformation}. From MIST, we use only Study 1 (MIST-1) which includes 409 participants, each providing judgment on the same 100 claims, and demographic information including age, gender, and education. 
From the two datasets combined, we obtain 13.8K claims for evaluation.


\subsection{Belief Data}
We collect two complementary belief signals, consisting of individually observed belief judgments and group-level (demographic) belief distributions:

\noindent \textbf{(1) Observed Data} - claims that were directly judged by participants in PANDORA and MIST-1 datasets. These responses capture individual-level belief judgments, reflecting each participant’s personal stance rather than demographic group averages. We use two claim judgments as observed beliefs for each participant (keeping it separate from the evaluation data). This leads to 27.6 claim judgments as observed beliefs.

\noindent \textbf{(2) Imputed Data} - inferred from the World Values Survey Wave 7\footnote{\url{https://www.worldvaluessurvey.org/WVSDocumentationWV7.jsp}} distributions. Imputed data represent demographic belief priors (group–level), inferred from WVS, conditioned solely on demographic attributes. We map these imputed belief items to our belief taxonomy using exploratory factor analysis. This yields 126 imputed belief questions and corresponding demographic distributions. Table~\ref{tab:belief_taxonomy} shows examples, and Appendix~\ref{sec:wvs} contains mapping details and all questions.

\definecolor{CustomPastelPink}{RGB}{255, 120, 200}
\definecolor{CustomGreen}{RGB}{0, 180, 150}
\definecolor{CustomBlue}{RGB}{0, 100, 200}
\definecolor{CustomPastelYellow}{RGB}{253, 253, 150}

\begin{figure}[t]
\begin{tcolorbox}[
    enhanced,
    colback=CustomGreen!20, 
    colframe=CustomGreen!80,
    title= Dataset Examples,
    fonttitle=\bfseries,
    boxrule=0.8pt
]

\textbf{\underline{MIST:}} \\
\textbf{Participant ID}: 1 \\
\textbf{Participant Demographic Group}: d1 \\
\textbf{Claim}: The Government Is Knowingly Spreading Disease Through the Airwaves and Food Supply \\
\textbf{Participant Choice}: fake \\
\textbf{Gold Label}: fake \\

\medskip

\textbf{\underline{PANDORA:}} \\
\textbf{Participant ID}: 1 \\
\textbf{Participant Demographic Group}: d1 \\
\textbf{Claim}: News: Updated CDC Covid Numbers Show Only Six Percent of Total US Deaths Actually Dies Solely From Virus. \\
\textbf{Participant Choice}: true \\
\textbf{Gold Label}: fake \\

\end{tcolorbox}
\caption{Dataset Examples - MIST and PANDORA}
\label{fig:bench}
\end{figure}

Fig~\ref{fig:bench} shows an example from each MIST and PANDORA datasets. Both datasets are similar, except for the size. Furthermore, MIST does not contain the living area information of the participants, therefore, we do not use it for fine-tuning purposes.

\section{Prompt-Based Conditioning}
\label{sec:promptbased}

\subsection{Prompt Details}

\begin{figure}[t]
\begin{tcolorbox}[
    enhanced,
    colback=CustomBlue!10, 
    colframe=CustomBlue!80,
    title= Prompt,
    fonttitle=\bfseries,
    boxrule=0.8pt
]

You are a persona grounded by attributes: <demographic group>.  \\

Past beliefs and priors for this persona (for context, do not re-evaluate them): <beliefs>. \\

When judging a claim, stay consistent with this persona's prior beliefs where reasonable.

\medskip

\end{tcolorbox}
\caption{Prompt Conditioning Experiments}
\label{fig:bench}
\end{figure}

\begin{figure}
\centering
\includegraphics[width=\linewidth]{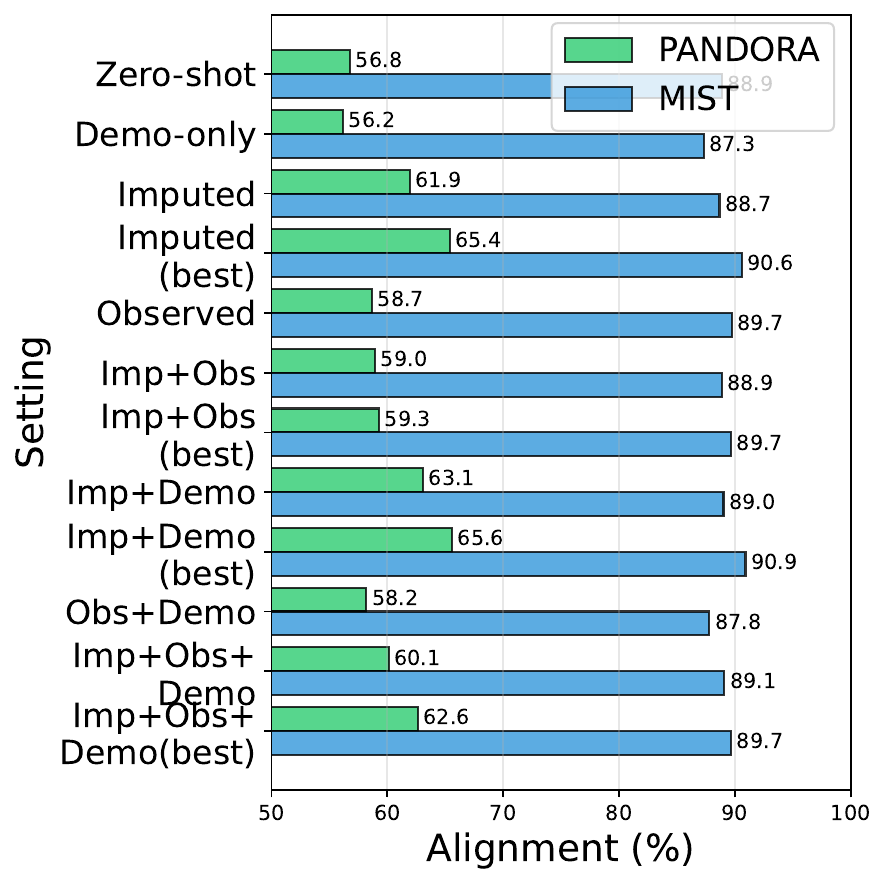}
\caption{Susceptibility Alignment across settings and datasets}
\label{fig:overall_comp_all}
\end{figure}

Fig~\ref{fig:overall_comp_all} shows the susceptibility alignment across the belief and demographic configurations for both datasets. For \texttt{Imputed} cases, we average across all belief dimensions, and \texttt{Imputed (best)} cases, we select the belief dimension that achieves the highest susceptibility alignment. Findings show that incorporating belief information consistently improves performance over the zero-shot and demographic-only baselines. We next analyze the impact of specific belief types, demographic ablations, datasets, and model performances.

\subsection{Belief-Dimension Analysis}
\label{sec:belief_ana}
Table~\ref{tab:belief_dimension} shows the belief-dimension-wise scores for each dataset and demographic dimension. 

Across both datasets, single dimensions consistently outperform the combined prior, and this is especially stark on PANDORA, where all-dimensions collapses to much lower accuracy while the best single dimensions remain high. 

For Prolific, the strongest modal scores are generally obtained from conspiracy mindset and epistemic/trust beliefs, especially for gender, age, and education. For MIST, cognitive-style beliefs and conspiracy mindset are the strongest individual dimensions, with cognitive-style beliefs reaching the highest education-based score and conspiracy mindset performing best for age. The all-dimensions setting is not uniformly strongest, suggesting that compact dimension-specific belief priors can be more informative than combining all belief dimensions at once. Rural/urban results are omitted in the updated modal table because the corresponding screened runs were unavailable or excluded.


  \begin{table*}[h]
  \centering
  \resizebox{\textwidth}{!}{%
  \begin{tabular}{llcccc}
  \hline
  \textbf{Dataset} & \textbf{Dimension} & \textbf{Gender} & \textbf{Age} & \textbf{Education} & \textbf{Rural/Urban} \\
  \hline
  \multirow{8}{*}{\textbf{Prolific-dataset}}
  & worldview\_and\_identity & 55.69 & 55.88 & 55.51 & 55.80 \\
  & epistmetic\_and\_trust\_beliefs & 58.23 & 58.82 & 58.90 & 58.65 \\
  & cognitive\_style\_beliefs & 55.24 & 57.35 & 55.72 & 56.10 \\
  & conspiracy\_mindset & 59.13 & 58.24 & 59.53 & 58.90 \\
  & moral\_and\_value\_beliefs & 55.99 & 55.00 & 54.66 & 55.20\\
  & emotion\_related & 55.39 & 58.24 & 54.66 & 56.10 \\
  & heurisitics & 54.94 & 59.41 & 54.24 & 56.53 \\
  & alldimensions & 57.12 & 58.24 & 56.67 & 57.50 \\
  \hline
  \multirow{8}{*}{\textbf{MIST-dataset}}
  & worldview\_and\_identity & 75.99 & 76.17 & 76.23 & \\
  & epistmetic\_and\_trust\_beliefs & 74.84 & 73.61 & 76.10 &\\
  & cognitive\_style\_beliefs & 79.48 & 78.81 & 81.17 & \\
  & conspiracy\_mindset & 77.85 & 80.39 & 79.17 & \\
  & moral\_and\_value\_beliefs & 74.05 & 75.29 & 73.91 &  \\
  & emotion\_related & 75.14 & 72.00 & 72.82 & \\
  & heurisitics & 75.89 & 76.29 & 75.93 &  \\
  & alldimensions & 86.00 & 74.50 & 73.86 &\\
  \hline
  \end{tabular}%
  }
  \caption{Belief Dimension Analysis using modal runs. }
  \label{tab:belief_dimension}
  \end{table*}


\begin{table*}
\centering
\small
\begin{tabular}{@{}lllp{6.5cm}@{}}
\toprule
\textbf{Demographic} & \textbf{Model} & \textbf{Topic/Theme} & \textbf{Effect and Notes} \\
\midrule
\multirow{5}{*}{\textbf{Age}} 
& Llama-3-8B & New study \& ideology (T1) & +2.8 ppts (60+--$<30$). Older: 0.984 vs younger: 0.956. \\
& Llama-3-8B & Global threats \& population (T2) & +2.0 ppts. Older: 0.961 vs younger: 0.941. \\
& Llama-3-8B & Marijuana/new-study (T4) & +1.7 ppts. 60+ group: 0.947 vs <30: 0.930. \\
& Qwen-2.5-14B & Marijuana/new-study (T4) & +2.5 ppts. 60+ group: 0.917 vs <30: 0.892. \\
& Mistral-7B & All topics & $\leq$0.5 ppts. Age differences negligible. \\
\midrule
\multirow{5}{*}{\textbf{Education}}
& Llama-3-8B & Sleep/blue-light health (T2) & +8.4 ppts. Completed: 0.918 vs not: 0.833 (six cases). \\
& Qwen-2.5-14B & Politics/elections (T3) & +16.7 ppts. Completed perfect; not-completed: 0.833 (six cases). \\
& Llama-3-8B & Marijuana/new-study (T4) & +0.9 ppts. Difference: 0.962 vs 0.953. \\
& Mistral-7B & Marijuana/new-study (T4) & +0.2 ppts. Effect negligible. \\
& All models & Other topics & $\leq$0.0–0.2 ppts. No meaningful education effect. \\
\midrule
\multirow{6}{*}{\textbf{Gender}}
& Llama-3-8B & Global influence/future threats (T0) & $\leq$ 2 ppts ($|F-M|$). Nearly identical accuracies. \\
& Llama-3-8B & Science \& health claims (T2) & $\leq$ 2 ppts. No consistent female–male separation. \\
& Llama-3-8B & Public opinion/social beliefs (T4) & $\approx$ 0 ppts. Near-ceiling for both genders. \\
& Mistral-7B & Global influence/future threats (T0) & $\approx$ 0 ppts. Accuracies overlap almost exactly. \\
& Mistral-7B & Marijuana/public opinion (T4) & $\leq$ 1 ppt. No directional trend. \\
& Qwen-2.5-14B & All topics & 0 ppts. Identical accuracies across all topics. \\
\bottomrule
\end{tabular}
\caption{Thematic Analysis across Demographics.}
\label{tab:demographics}
\end{table*}

  \begin{table*}[ht]
  \centering
  \scriptsize
  \setlength{\tabcolsep}{3pt}
  \resizebox{\textwidth}{!}{%
  \begin{tabular}{lllcccccc}
  \toprule
  \textbf{Data} & \textbf{Model} & \textbf{Belief Type} & \textbf{Zero-shot} & \textbf{Demo-only} & \textbf{Imputed} &
  \textbf{Observed} & \textbf{Imp+Obs} & \textbf{Imp+Demo} \\
  \midrule
  PANDORA & Qwen2.5-14B & modal & 62.28 & 62.28 & 64.51 & 62.08 & 63.49 & \textbf{64.81} \\
  PANDORA & Qwen2.5-14B & dist  & 62.28 & 62.28 & 63.51 & 62.08 & 62.79 & \textbf{63.91} \\
  PANDORA & Llama-3-8B & modal & 57.49 & 56.89 & 60.63 & 58.98 & 60.01 & \textbf{60.93} \\
  PANDORA & Llama-3-8B & dist  & 57.49 & 56.89 & 59.63 & 58.98 & 59.31 & \textbf{60.03} \\
  PANDORA & Llama-3.2-3B & modal & 53.89 & 57.49 & 56.76 & \textbf{60.78} & 58.81 & 55.96 \\
  PANDORA & Llama-3.2-3B & dist  & 53.89 & 57.49 & 60.43 & \textbf{60.78} & 59.77 & 58.74 \\
  PANDORA & Grok-4-Fast & modal & 57.07 & 56.60 & 58.20 & 57.00 & 57.85 & \textbf{58.55} \\
  PANDORA & Grok-4-Fast & dist  & 57.07 & 56.60 & 57.65 & 57.00 & 57.40 & \textbf{57.95} \\
  PANDORA & Mistral-7B-v0.2 & modal & 55.99 & 54.19 & 55.82 & 55.79 & 55.82 & \textbf{56.12} \\
  PANDORA & Mistral-7B-v0.2 & dist  & \textbf{55.99} & 54.19 & 54.82 & 55.79 & 55.30 & 55.22 \\
  PANDORA & OLMo-2-7B & modal & 50.60 & 49.40 & 49.43 & \textbf{58.38} & 50.51 & 50.23 \\
  PANDORA & OLMo-2-7B & dist  & 50.60 & 49.40 & 52.14 & \textbf{58.38} & 52.57 & 52.54 \\
  PANDORA & DeepSeek-7B & modal & 52.40 & 50.00 & 52.53 & 41.32 & 48.96 & \textbf{53.33} \\
  PANDORA & DeepSeek-7B & dist  & 52.40 & 50.00 & 52.85 & 41.32 & 48.04 & \textbf{53.25} \\
  \midrule
  MIST & Qwen2.5-14B & modal & 95.97 & 95.97 & 97.41 & 95.77 & 96.59 & \textbf{97.72} \\
  MIST & Qwen2.5-14B & dist  & 95.97 & 95.97 & 96.98 & 95.77 & 96.37 & \textbf{97.18} \\
  MIST & Llama-3-8B & modal & \textbf{93.32} & 91.70 & 92.51 & 86.56 & 92.51 & 93.00 \\
  MIST & Llama-3-8B & dist  & 93.32 & 91.70 & 91.51 & 86.56 & \textbf{93.59} & 92.10 \\
  MIST & Llama-3.2-3B & modal & 91.08 & 88.96 & 87.13 & 76.82 & \textbf{94.13} & 87.49 \\
  MIST & Llama-3.2-3B & dist  & 91.08 & 88.96 & 88.69 & 76.82 & \textbf{93.87} & 88.29 \\
  MIST & Grok-4-Fast & modal & 87.20 & 86.70 & 88.30 & 87.05 & 87.95 & \textbf{88.60} \\
  MIST & Mistral-7B-v0.2 & modal & 71.91 & 68.23 & 73.67 & 71.71 & 72.89 & \textbf{73.97} \\
  MIST & Mistral-7B-v0.2 & dist  & 71.91 & 68.23 & 72.67 & 71.71 & 72.19 & \textbf{73.07} \\
  MIST & OLMo-2-7B & modal & 79.83 & 81.49 & 77.35 & 80.83 & 80.15 & \textbf{82.29} \\
  MIST & OLMo-2-7B & dist  & 79.83 & 81.49 & 80.38 & 80.83 & \textbf{82.23} & 81.89 \\
  MIST & DeepSeek-7B & modal & 63.97 & 64.64 & 63.61 & 56.46 & 62.16 & \textbf{65.44} \\
  MIST & DeepSeek-7B & dist  & 63.97 & 64.64 & 62.55 & 56.46 & 64.43 & \textbf{65.04} \\
  \bottomrule
  \end{tabular}%
  }
  \caption{Ablations comparing modal belief conditioning and belief-distribution conditioning across datasets and models.
  Scores are averaged over demographic axes. Best value per row is in bold.}
  \label{tab:distribution_ablation}
  \end{table*}


\subsection{Dataset differences - PANDORA vs MIST}

Lower accuracies on PANDORA are expected because Prolific is currently much smaller than MIST, even though both evaluate individual-level beliefs. The smaller Prolific set has less coverage across claims, demographics, and belief patterns, making estimates more sensitive to participant-level variation and noise. In contrast, the larger MIST dataset provides broader coverage and more stable training/evaluation signals, which can lead to higher accuracies.

\subsection{Thematic Analysis}
\label{sec:thematic}

We perform topic clustering on the news claims to identify latent topical grouping and investigate whether LLM susceptibility alignment vary across demographics within each topic. We do this for three models -- Llama-3-8b, Mistral-7b and Qwen-2.5-14b. We perform this on Dataset 2, as it is much larger and diverse than Dataset 1. Using non-negative matrix factorization on claim texts, we identify five dominant topics: (1) global power, population, and future threats; (2) political leaders and historical rankings; (3) science, health, and technology claims; (4) politics, government, and information-control narratives; and (5) public opinion, moral values, and social belief statements. Topic-level analysis shows that susceptibility alignment is not uniform across demographics. Topics related to political rankings and public opinion (Topics 2 and 5) achieve uniformly high accuracy across all demographic groups, resulting in negligible age, gender, or education gaps. Contrastingly, science/health claims and government-related narratives (Topics 3 and 4) exhibit the largest demographic variation, with older and higher-education groups generally achieving higher accuracy. Gender-based differences are minimal across all topics. Table~\ref{tab:demographics} shows results across some topics and models showing differences across demographics. 

These findings suggest that demographic susceptibility in LLMs is not just a broad, identity-driven phenomenon but also contextual and topic-sensitive. This also  aligns with interdisciplinary work in psychology and communication suggesting that demographic susceptibility to misinformation is highly context-dependent. Prior studies show that gender alone is a weak predictor of misinformation belief, with analytic thinking and reasoning style playing a much larger role~\cite{marinescu2018quasi}. In contrast, age and education demographics differ more for complex or ambiguous claims, particularly in science and policy domains, consistent with work on cognitive processing, and motivated reasoning~\cite{kahan2017misconceptions}. From a cognitive perspective, this pattern supports dual-process theories of reasoning, whereby strong lexical reduce individual differences, while inference-heavy claims amplify them~\cite{evans2013dual}. 

\subsection{Experiments using belief-distributions}
\label{sec:dis_prompt}

In our main experiments, we utilize modal responses for prompt conditioning. Here, we perform additional experiments that use belief
distributions instead of modal responses in prompts. Table~\ref{tab:distribution_ablation} compares modal and
distribution-based prompting across models on PANDORA and MIST.

Overall, distribution-based conditioning is competitive with modal prompting, but does not consistently outperform it. Across paired comparisons, distributional prompts sometimes improve imputed-only or imputed+observed settings, suggesting that population-level uncertainty can provide useful signal. However, these gains are model- and dataset-dependent. In particular, modal prompting is more stable when imputed beliefs are combined with demographics, which is our strongest setting in the main experiments. This pattern suggests that full distributions may introduce additional ambiguity: they encode group-level uncertainty, however, the prediction target is a binary individual judgment. As a result, models may struggle to decide whether to follow the majority response, sample from the distribution, or treat minority probabilities as competing evidence.

We therefore use modal belief responses in the main experiments for three reasons: (1) they provide a simpler and more interpretable representation of demographic belief priors. (2) they reduce prompt length and avoid requiring the model to perform implicit probabilistic reasoning over Likert distributions. (3) they lead to more stable performance in the main combined settings, especially when beliefs are paired with demographic information. Thus, while distributional prompts are a useful robustness check, modal responses provide a cleaner and more reliable prompt representation for our primary experiments given the models.


\subsection{Distributional alignment within groups}
\label{app:distribution_alignment}
We use Jensen--Shannon (JS) divergence across demographic groups to investigate how models can capture the distributions in demographic groups. Distributional representations indicate how much uncertainty remains within each group, while modal representations show what happens when this uncertainty is collapsed to a single dominant response. We also report modal disagreement, the percentage of WVS items for which demographic groups have different modal responses.

\begin{table}[ht]
\centering
\small
\begin{tabular}{lccc}
\toprule
\textbf{Axis} & \textbf{Dist. JS} & \textbf{Modal JS} & \textbf{Modal Disagree.} \\
\midrule
Age & 0.0421 & 0.2511 & 25.53 \\
Education & 0.0498 & 0.2454 & 24.96 \\
Gender & 0.0083 & 0.1175 & 11.95 \\
Rural/Urban & 0.0072 & 0.0958 & 9.76 \\
\bottomrule
\end{tabular}
\caption{Average demographic divergence for distribution-based and modal belief prompt representations. JS divergence is reported in bits; modal disagreement reports the percentage of WVS items for which demographic groups have different modal responses.}
\label{tab:modal_vs_distribution_js}
\end{table}

Table~\ref{tab:modal_vs_distribution_js} shows that full distributional representations retain substantially more uncertainty, producing lower JS divergence across demographic groups. In contrast, modal prompts collapse uncertainty and produce sharper demographic separation. This does not mean modal prompts are always more faithful to the underlying population; rather, they provide a compact approximation of the dominant belief signal. The results therefore highlight a trade-off: modal prompts are simpler and more stable for prompt-conditioning, while distributional representations better preserve within-group uncertainty. This motivates our additional distributional prompting experiments and our use of belief distributions in BeliefSim-FT, where uncertainty can be represented more directly through distribution-level training.

\subsection{Evaluation Under Human Judgment Variability}
\label{app:human_variability}
Our main evaluation metric treats each participant judgment as the prediction target, reflecting our goal of simulating individual-level susceptibility rather than estimating a single claim-level consensus label. However, susceptibility judgments are inherently noisy: participants may disagree on the same claim, and some claims may have higher response entropy than others. We therefore interpret susceptibility alignment as agreement with observed individual responses, not as recovery of an objective or consensus truth label.

\begin{table}[!ht] \centering \small \begin{tabular}{lccc} \toprule \textbf{Dataset} & \textbf{Low} & \textbf{Mid} & \textbf{High} \\ \midrule PANDORA & 58.24 & 51.03 & 42.95 \\ MIST & 86.13 & 78.13 & 87.77 \\ 
\bottomrule 
\end{tabular} 
\caption{Modal prompt-conditioning accuracy by claim-level entropy.} 
\label{tab:entropy_dataset} 
\end{table}

To better contextualize model performance, we examine whether accuracy varies with claim-level entropy, estimated from the variability of human responses to each claim (Table~\ref{tab:entropy_dataset}). Higher entropy indicates greater participant disagreement and therefore a noisier prediction target. For PANDORA, accuracy decreases from 58.2\% on low-entropy claims to 42.9\% on high-entropy claims, suggesting that ambiguity in participant judgments partly explains model errors. MIST remains comparatively stable across entropy bins, with only a mid-entropy dip. This distinction is important because it separates model limitations from intrinsically noisy human targets. The results suggest that ambiguity partly explains PANDORA errors, whereas screened MIST performance remains comparatively stable across agreement levels.







\subsection{Veracity Prediction vs. Susceptibility Prediction}
\label{app:veracity_vs_susceptibility}

We additionally compare zero-shot veracity accuracy with belief-conditioned susceptibility alignment (best setting). Veracity accuracy measures whether the model prediction matches the gold truth label of a claim, while susceptibility alignment measures whether the model prediction matches a participant's observed judgment. These are related but distinct tasks: a model can be factually correct while failing to simulate whether a participant believes the claim.

Table~\ref{tab:veracity_vs_susceptibility} compares two distinct evaluation targets: veracity prediction, which measures agreement with the gold truth label, and susceptibility prediction, which measures agreement with participants' observed judgments. The results show that these objectives do not always move together. For several open-source models, belief-conditioned susceptibility alignment is higher than zero-shot veracity accuracy, especially on PANDORA, suggesting that belief conditioning can help approximate human judgments beyond factual correctness alone. However, the pattern is not uniform: \texttt{Grok-4-Fast} has strong veracity accuracy but lower susceptibility alignment, while some MIST models show only small differences between the two metrics.

Overall, this comparison highlights that factual veracity correctness and human-belief simulation capture different aspects of misinformation response. Veracity prediction evaluates whether a model identifies the correct truth status of a claim, while susceptibility prediction evaluates whether it can approximate how a specific participant judges that claim. Thus, high performance on one objective does not necessarily imply high performance on the other.

  \begin{table*}[t]
  \centering
  \small
  \setlength{\tabcolsep}{5pt}
  \begin{tabular}{llccc}
  \toprule
  \textbf{Dataset} & \textbf{Model} & \textbf{Veracity} & \textbf{Belief-Susc.} & \textbf{$\Delta$} \\
  \midrule
  PANDORA & Qwen2.5-14B & 62.28 & 64.81 & +2.53 \\
  PANDORA & Llama-3-8B & 57.49 & 60.93 & +3.44 \\
  PANDORA & Llama-3.2-3B & 53.89 & 60.78 & +6.89 \\
  PANDORA & Grok-4-Fast & 66.77 & 58.55 & -8.22 \\
  PANDORA & Mistral-7B-v0.2 & 55.99 & 56.12 & +0.13 \\
  PANDORA & OLMo-2-7B & 50.60 & 58.38 & +7.78 \\
  PANDORA & DeepSeek-7B & 52.40 & 53.33 & +0.93 \\
  \midrule
  MIST & Qwen2.5-14B & 96.97 & 97.72 & +0.75 \\
  MIST & Llama-3-8B & 93.32 & 93.00 & -0.32 \\
  MIST & Llama-3.2-3B & 91.08 & 94.13 & +3.05 \\
  MIST & Grok-4-Fast & 97.70 & 88.60 & -9.10 \\
  MIST & Mistral-7B-v0.2 & 71.91 & 73.97 & +2.06 \\
  MIST & OLMo-2-7B & 79.83 & 82.29 & +2.46 \\
  MIST & DeepSeek-7B & 63.97 & 65.44 & +1.47 \\
  \bottomrule
  \end{tabular}
  \caption{Comparison between zero-shot veracity accuracy and belief-conditioned susceptibility alignment. Veracity measures
  agreement with the gold truth label, while belief-susceptibility measures agreement with human judgments under belief-
  conditioned prompting.}
  \label{tab:veracity_vs_susceptibility}
  \end{table*}

\subsection{Demographic Counterfactual Analysis}
\label{app:flip_rates}

Here, we provide flip rates across the three metrics in Table~\ref{tab:counterfactual_by_demo}. Overall, \texttt{Mistral} shows the lowest flip rates, suggesting limited dependence on demographic cues. \texttt{Qwen} is also comparatively stable and low. In contrast, \texttt{Llama} and \texttt{Deepseek} models show substantially higher demographic sensitivity. Across models, the largest flip rates concentrate in Education and Living Area, especially in the Shortcut Reliance setting, where demographics are designed to be non-predictive. This could also relate to stereotype-driven reliance.

\begin{table*}[ht]
\centering
\small
\resizebox{\textwidth}{!}{
\begin{tabular}{lrrrrrrrrrrrr}
\toprule
\multirow{2}{*}{Model}
& \multicolumn{4}{c}{Panel A: Utility}
& \multicolumn{4}{c}{Panel B: Shortcut}
& \multicolumn{4}{c}{Panel C: Complementarity} \\
\cmidrule(lr){2-5}\cmidrule(lr){6-9}\cmidrule(lr){10-13}
& Gender & Area & Age & Edu.
& Gender & Area & Age & Edu.
& Gender & Area & Age & Edu. \\
\midrule
Mistral-7B-v0.2
& 0.13 & 0.67 & 0.07 & 0.64
& 2.38 & 0.00 & 2.78 & 0.00
& 3.11 & 0.35 & 3.25 & 2.39 \\

Qwen3-4B
& 0.40 & 0.67 & 0.59 & 2.41
& 0.60 & 1.25 & 2.78 & 0.00
& 1.56 & 12.32 & 7.02 & 5.38 \\

OLMo-2-7B
& 2.00 & 7.67 & 1.45 & 2.20
& 7.74 & 12.66 & 0.00 & 0.00
& 5.20 & 9.51 & 11.34 & 10.76 \\

Qwen2.5-14B
& 4.41 & 2.33 & 0.72 & 1.99
& 2.98 & 7.50 & 0.00 & 0.00
& 5.02 & 4.93 & 2.91 & 3.59 \\

Llama-3.2-3B
& 10.28 & 6.00 & 12.15 & 12.78
& 8.33 & 5.00 & 2.78 & 16.67
& 2.42 & 9.64 & 3.81 & 5.62 \\

DeepSeek-7B
& 5.87 & 13.00 & 9.65 & 22.87
& 7.14 & 17.50 & 2.78 & 25.00
& 2.60 & 8.45 & 10.79 & 11.75 \\

Llama-3-8B
& 1.80 & 8.00 & 10.51 & 32.46
& 8.33 & 6.25 & 13.89 & 33.33
& 6.06 & 8.45 & 3.25 & 5.38 \\
\bottomrule
\end{tabular}
}
\caption{Counterfactual flip rates (\%) across demographic groups. Panel A measures demographic-swap sensitivity on the
original distribution, Panel B measures shortcut reliance on balanced slices, and Panel C measures complementarity under
degraded beliefs.}
\label{tab:counterfactual_by_demo}
\end{table*}

\begin{table}[!ht]
\centering
\small
\begin{tabular}{lrrr}
\toprule
\textbf{Condition} & \textbf{Llama} & \textbf{Qwen} & \textbf{Mistral} \\
\midrule
Zero-shot                & $0.115$ & $0.107$ & $0.081$ \\
Demo                     & $0.042$ & $0.107$ & $0.028$ \\
Imputed                  & $-0.023$ & $0.107$ & $0.028$ \\
Imputed + Demo           & $-0.023$ & $0.107$ & $0.028$ \\
\midrule
Observed                 & $-0.025$ & $0.031$ & $0.034$ \\
Observed + Demo          & $-0.023$ & $0.036$ & $0.034$ \\
Imp.\ + Obs.\            & $-0.020$ & $0.032$ & $0.034$ \\
Imp.\ + Obs.\ + Demo     & $-0.025$ & $0.058$ & $0.024$ \\
\bottomrule
\end{tabular}
\caption{Spearman $\rho$ between zero-shot factual confidence and human-label
match rate on MIST-1, restricted to false claims. Positive values indicate
that higher model confidence in a claim's falsity correlates with better
matching of human judgments overall; near-zero or negative values indicate
decoupling from the factual prior.}
\label{tab:conf_corr}
\end{table}

\begin{table*}[ht]
\centering
\small
\begin{tabular}{llclp{6.5cm}}
\toprule
Dataset & Model & Axis & Hit rows & Representative raw-output evidence \\
\midrule
MIST-1 & Mistral-7B-Instruct-v0.2 & Education & 36/36
& ``A person with a \textbf{completed education is less likely to believe in such a sensational and unsubstantiated claim}.'' \\

MIST-1 & Mistral-7B-Instruct-v0.2 & Education & 36/36
& ``This person, with \textbf{not completed education}, is \textbf{less likely to have the knowledge or ability to make an informed
judgment}.'' \\

MIST-1 & Mistral-7B-Instruct-v0.2 & Gender & 30/54
& ``This person, being \textbf{male}, is likely to believe headlines that discuss \textbf{international politics and economic
transactions.}'' \\

MIST-1 & Mistral-7B-Instruct-v0.2 & Gender & 29/54
& ``This person, a \textbf{female}, is likely to be informed about \textbf{global news and economics.}'' \\

PANDORA & OLMo-2-1124-7B-Instruct & Living Area & 2/80
& ``The person living in a \textbf{rural area} might be more inclined to trust the news due to their \textbf{exposure to fewer alternative
sources}.'' \\

PANDORA & Llama-3.2-3B-Instruct & Living Area & 10/80
& ``The person's perspective is influenced by their \textbf{living area, which is rural,} and may \textbf{not be familiar with the specifics
of the Ferguson incident}.'' \\


\bottomrule
\end{tabular}
\caption{Qualitative examples of explicit demographic references in raw outputs for Panel B shortcut-reliance cases. Counts
indicate rows whose raw output mentions the corresponding demographic axis; excerpts are illustrative and are not used as
the primary metric.}
\label{tab:shortcut_raw_demo_reasoning}
\end{table*}

\noindent \textbf{Qualitative evidence of shortcut reasoning.}. To complement the flip-rate analysis, Table~\ref{tab:shortcut_raw_demo_reasoning} shows examples where model outputs explicitly reference demographic attributes in Panel B, where demographics are non-predictive by construction. These examples illustrate that some models not only change predictions under demographic swaps, but also justify their decisions using demographic assumptions about education, gender, age, or living area. This supports our interpretation that high flip rates in the shortcut-reliance setting can reflect demographic shortcutting or stereotype-like reasoning. However, these excerpts are used only as qualitative evidence.

\subsection{Factual Confidence and Susceptibility Simulation}
\label{sec:factual-confidence}

A potential explanation for the limited gains from belief conditioning is that LLMs' pretraining knowledge about claim veracity acts as a compounding bias: when a model is confident a claim is false, it may resist simulating a persona that would nonetheless believe it. We analyze this hypothesis in this section.

\paragraph{Method.}
For each claim in MIST-1, we extract a zero-shot factual confidence score using a logit-based probe: we compute the renormalized probability mass assigned to \texttt{true} vs.\ \texttt{false} at the first generated token position, yielding a scalar in $[0.5, 1.0]$ indicating confidence in the predicted label. We then compute the Spearman rank correlation ($\rho$) between this confidence score and whether the model's predicted label matches the human's belief judgment, for each (model, condition) pair. We restrict the analysis to \textit{false claims only} (gold label: misinformation), since factual override is operative precisely when the model's factual prior conflicts with the susceptible human's belief.

\paragraph{Results.}
Table~\ref{tab:conf_corr} reports Spearman $\rho$ across all prompting conditions. On MIST-1, zero-shot correlations are small but consistently positive across all models (Llama: 0.115, Qwen: 0.107, Mistral: 0.081), confirming that factual confidence \textit{somewhat} biases models toward predicting \textit{fake}---which coincidentally aligns with the skeptical majority of annotators but at the cost of failing susceptible individuals. As belief conditioning is introduced, Llama and Mistral shift toward zero and slightly negative values under imputed conditions, suggesting partial decoupling from the factual prior. Qwen, however, remains at $\rho = 0.107$ across all modal conditions, consistent with its lower sensitivity to prompt-based conditioning observed in the counterfactual analysis (\S\ref{sec:counterfactual-subsec}). Observed belief conditions bring all models near zero, indicating that individual-level belief signal is most effective at overriding the factual prior.

Taken together, factual confidence introduces a modest positive bias in zero-shot settings but is not the dominant factor limiting susceptibility simulation. Belief conditioning---particularly with observed beliefs---largely neutralizes it, suggesting the broader simulation gap is driven by other mechanisms.

\section{Fine-tuning Analysis}

\label{sec:ft_ana}

\subsection{MIST-1/MIST-2b overlap discussion}
\label{app:mist_overlap}

Our fine-tuning experiments use MIST-1 for training and both MIST-1 and MIST-2b for evaluation. Since the MIST benchmark is a standardized psychometric instrument, MIST-2b contains items derived from the MIST-1 item-development pool. Therefore, the MIST-2b evaluation is not a claim-disjoint test of generalization to entirely unseen misinformation claims. Instead, it evaluates cross-study and cross-participant generalization under a shared measurement instrument. This distinction is important because, for our susceptibility simulation setting, the primary leakage risk is participant-level leakage: if the same participant appeared in both training and evaluation, the model could learn participant-specific response tendencies rather than generalizable belief-susceptibility mappings. We avoid this by using independent participant samples across MIST-1 and MIST-2b. Claim overlap is less central to this particular evaluation because the goal is to predict individual susceptibility responses under a validated fixed instrument, rather than to test open-domain claim generalization. We therefore report MIST-2b results as shared-instrument cross-participant transfer, and leave fully claim-disjoint evaluation as a stricter future setting.

\subsection{Phase 1 results}
For Phase 1, Table~\ref{tab:kl_divergence} shows that Qwen has the lowest KL divergence, followed by Mistral and Llama for both settings - head tuning and full-fine-tuning. With LoRA-ft, the trends remain similar but slightly lower than head-ft. This indicates that Qwen more faithfully captures population-level belief distributions, a critical property for reliable downstream simulation of demographic misinformation susceptibility. Other models also perform competitively, suggesting that most models LLMs can capture belief distributions to a reasonable extent.

\begin{table}[h]
\centering
\begin{tabular}{lcc}
\hline
\textbf{Model} & \textbf{Head Tuning} & \textbf{Full Fine-tuning} \\
\hline
Qwen & 0.0511 & 0.0732 \\
Mistral & 0.087 & 0.107 \\
Llama & 0.287 & 0.487 \\
\hline
\end{tabular}
\caption{KL divergence comparison across different models and tuning methods.}
\label{tab:kl_divergence}
\end{table}

\subsection{Phase 2 results}

\begin{table}[h]
\small
\centering
\resizebox{\columnwidth}{!}{%
\begin{tabular}{lccccc}
\toprule
\multirow{2}{*}{\textbf{Model}} & \multicolumn{2}{c}{\textbf{PANDORA+MIST-1}} & \multicolumn{2}{c}{\textbf{MIST-2}} \\
\cmidrule(lr){2-3} \cmidrule(lr){4-5}
& \textbf{Acc} & \textbf{F1} & \textbf{ Acc} & \textbf{F1} \\
\midrule
\multicolumn{5}{c}{\textbf{Baseline (1-phase)}} \\
\midrule
Llama & 0.750 & 0.751 & 0.680 & 0.673 \\
Qwen & 0.760 & 0.761 & 0.720 & 0.712 \\
Mistral & 0.740 & 0.761 & 0.650 & 0.653 \\
\midrule
\multicolumn{5}{c}{\textbf{LoRA-FT (2-phase)}} \\
\midrule
Llama & 0.809 & 0.809 & 0.876 & 0.861 \\
Qwen & 0.800 & 0.800 & 0.893 & 0.892 \\
Mistral & 0.800 & 0.790 & 0.887 & 0.867 \\
\midrule
\multicolumn{5}{c}{\textbf{BeliefSim-FT}} \\
\midrule
Llama & 0.793 & 0.793 & 0.884 & 0.896 \\
Qwen & 0.792 & 0.791 & 0.924 & 0.906 \\
Mistral & 0.792 & 0.792 & 0.884 & 0.823 \\

\bottomrule

\end{tabular}%
}
\caption{Fine-tuning performances}
\label{tab:ft2}
\end{table}

Table~\ref{tab:ft2} shows susceptibility alignment and F1 scores across different setups and models. Across methods, the key finding is that cross-study (MIST-2) generalization improves drastically once we move from 1-phase to 2-phase pipelines. The 1-phase baseline model perform decent, however, the performance drops sharply on MIST-2. For MIST-2, our BeliefSim-FT approach performs the best, showing that belief adapter + lightweight susceptibility head captures more transferable signals aligned with out-of-domain data. 

Comparing the 2-phase variants, we find that BeliefSim-FT remains the strongest on MIST-2, however, LoRA-ft slightly improves on PANDORA+MIST-1 but gives weaker transfer. This may be due to mild overfitting when more parameters are updated. Across models, Qwen consistently performs the best for 2-phase settings. Overall, the 2-phase training approach is helpful improving robustness and also adapting to out-of-domain data.

\subsection{Shortcut Reliance Experiments}
\label{sec:shortcut}
Table~\ref{tab:shortcut_reliance2} shows that shortcut reliance is low, especially under head tuning. We measure flip-rate: the fraction of evaluation examples for which the model’s predicted label changes when a shortcut feature is removed or perturbed, and Probability Delta: the average absolute change in the model’s predicted probability for the original class after the shortcut is removed.
On PROLIFIC+MIST-1, head-tuned models (Llama, Qwen and Mistral) show near-zero flip rates, indicating that removing the shortcut signal has little effect on predicted labels. Full fine-tuning also tends to have small flip rates and probability deltas in-domain, but they are more frequently non-zero (e.g., Llama/Mistral flip more than head tuning), consistent with the idea that updating more parameters can slightly increase sensitivity to spurious cues.

On MIST-2, the most notable shortcut sensitivity appears for Qwen under full fine-tuning (flip rate 0.1098, larger prob delta), whereas the corresponding head-tuned Qwen model shows 0 flip rate. This pattern supports the our previous findings that two-phase-head-only training is more robust: it preserves performance while reducing reliance on shortcut features that do not transfer.

\begin{table*}[h]
\centering
\small
\begin{tabular}{llccc|ccc}
\hline
\multirow{2}{*}{\textbf{Method}} & \multirow{2}{*}{\textbf{Model}} & 
\multicolumn{3}{c|}{\textbf{PANDORA+MIST-1}} & 
\multicolumn{3}{c}{\textbf{MIST-2}} \\
\cline{3-8}
& & \textbf{Flip Rate} & \textbf{Prob Delta} & \textbf{Acc Drop} 
  & \textbf{Flip Rate} & \textbf{Prob Delta} & \textbf{Acc Drop} \\
\hline
\multirow{3}{*}{Full-FT} 
& Llama   & 0.0239 & 0.0148 & 0.0049 & 0.0000 & 0.0000 & 0.0000 \\
& Qwen    & 0.0173 & 0.0476 & 0.0012 & 0.1098 & 0.0700 & 0.1027 \\
& Mistral & 0.0323 & 0.0024 & 0.0000 & 0.0000 & 0.0129 & 0.0000 \\
\hline
\multirow{3}{*}{Head Tuning} 
& Llama   & 0.0000 & 0.0125 & 0.0000 & 0.0000 & 0.0129 & 0.0000 \\
& Qwen    & 0.0080 & 0.0261 & 0.0012 & 0.0000 & 0.0427 & 0.0000 \\
& Mistral & 0.0000 & 0.0071 & 0.0000 & 0.0000 & 0.0069 & 0.0000 \\
\hline
\end{tabular}
\caption{Shortcut reliance comparison between Full Fine-tuning and Head Tuning across different models on PANDORA+MIST-1 and MIST-2 datasets.}
\label{tab:shortcut_reliance2}
\end{table*}

\subsection{Training Details}

\begin{figure}[t]
\begin{tcolorbox}[
    enhanced,
    colback=CustomBlue!10, 
    colframe=CustomBlue!80,
    title= Belief Modeling Prompt,
    fonttitle=\bfseries,
    boxrule=0.8pt
]

You are modeling belief distributions for a single demographic group. \\

Demographic : d1 \\

Question: q1 \\

There are k possible response options, numbered 1 through k.
Predict the probability for each option.

\end{tcolorbox}
\caption{Prompt for Phase-1 Belief Modeling}
\label{fig:bench}
\end{figure}

\subsubsection{2-phase Head-FT details}
Phase 1 trains a belief adapter to predict demographic-conditioned WVS response distributions. The model is a frozen encoder with a trainable linear head mapping the last-token hidden state to 10 logits, followed by softmax. Training uses AdamW (lr $5 \times 10^{-4}$, batch size 16, 2 epochs) and optimizes a scale-aware KL divergence loss, computed only over the valid bins 1..K for each example. For evaluation, we test generalization to unseen belief items, reporting KL distributional fit and majority-category accuracy.

Phase 2 trains a susceptibility prediction head on top of a frozen belief adapter learned in Phase 1. The model combines a frozen base encoder and frozen belief head to produce both a semantic representation of the input and a demographic-conditioned belief probability vector. These two are concatenated and fed into a lightweight, trainable classification head. We format the inputs as instruction-following prompts that include a demographic persona and up to two belief examples. During training, only the susceptibility head parameters are updated using cross-entropy loss, while the base model and belief adapter remain frozen. The model is optimized with AdamW (learning rate ($5 \times 10^{-4}$), batch size 8) for 2 epochs, with reproducible initialization via fixed random seeds. We finally evaluate at each epoch using accuracy and macro-F1.

\subsubsection{2-phase LoRA-FT details}
Here, we use Low-Rank Adaptation (LoRA) for parameter-efficient fine-tuning in both phases. In Phase 1, LoRA enables the model to adapt its representations to capture demographic belief distributions. In Phase 2, LoRA is again applied to the base model during susceptibility training, allowing the model to adjust task-relevant representations while keeping the number of trainable parameters small. 

\begin{table*}[h]
\centering
\tiny
\caption{World Values Survey Questions mapped to Misinformation Taxonomy Dimensions}
\label{tab:wvs_questions_mapped}
\begin{tabular}{p{3cm}p{1cm}p{11.5cm}}
\toprule
\textbf{Dimension} & \textbf{Q ID} & \textbf{Question} \\
\midrule
\multicolumn{3}{l}{\textbf{WorldView and Identity Beliefs}} \\
& Q6 & Importance of religion (1=Very important to 4=Not at all important) \\
& Q4 & Importance of politics (1=Very important to 4=Not at all important) \\
& Q171 & Frequency of religious service attendance (1=More than once/week to 7=Never) \\
& Q173 & Religious self-identification (1=Religious person, 2=Not religious, 3=Atheist) \\
& Q240 & Political left-right scale (1-10) \\
& Q254 & National pride (1=Very proud to 4=Not at all proud, 5=Not from US) \\
& Q255-Q259 & Closeness to village/region/country/continent/world (1=Very close to 4=Not close at all) \\
& Q19, Q23 & Unwanted neighbors: different race, different religion (1=Don't want, 2=Want) \\
& Q235-Q239 & Views on government styles: strong leader, technocracy, army rule, democracy, religious law (1=Very good to 4=Very bad) \\
& Q241-Q249 & Essential characteristics of democracy: tax/subsidize, religious law interpretation, free elections, unemployment aid, army takeover, civil rights, income equality, obedience, women's rights (1-10 scale) \\
\midrule
\multicolumn{3}{l}{\textbf{Epistemic and Trust Beliefs}} \\
& Q57 & General trust in people (1=Most can be trusted, 2=Need to be careful) \\
& Q58-Q63 & Trust in family, neighborhood, people you know, first-time meetings, other religions, other nationalities (1=Trust completely to 4=Don't trust at all) \\
& Q69, Q71, Q75 & Confidence in police, government, universities (1=A great deal to 4=None at all) \\
& Q158-Q163 & Views on science and technology: health/comfort, opportunities, faith vs science, moral breakdown, daily relevance, world better/worse (1-10 scale) \\
\midrule
\multicolumn{3}{l}{\textbf{Cognitive Style Beliefs}} \\
& Q152-Q153 & Priority ranking: economic growth, defense forces, participation in decisions, beautification (select most and next most important) \\
& Q154-Q155 & Priority ranking: order, participation, fighting prices, free speech (select most and next most important) \\
& Q156-Q157 & Priority ranking: stable economy, humane society, ideas over money, fight crime (select most and next most important) \\
& Q176 & Moral uncertainty: trouble deciding which moral rules are right (1-10 scale) \\
\midrule
\multicolumn{3}{l}{\textbf{Conspiracy Mindset}} \\
& Q112 & Perceived corruption level in country (1=No corruption to 10=Abundant corruption) \\
& Q113-Q117 & Corruption beliefs about state authorities, business executives, local authorities, civil service providers, journalists/media (1=None to 4=All of them) \\
& Q118 & Frequency of bribery needed for services (1=Never to 4=Always) \\
\midrule
\multicolumn{3}{l}{\textbf{Moral and Value Beliefs}} \\
& Q176 & Moral uncertainty (1-10 scale) \\
& Q177-Q195 & Justifiability of: claiming unentitled benefits, fare evasion, stealing, tax cheating, bribery, homosexuality, prostitution, abortion, divorce, premarital sex, suicide, euthanasia, wife beating, child beating, violence, terrorism, casual sex, political violence, death penalty (1=Never justifiable to 10=Always justifiable) \\
\midrule
\multicolumn{3}{l}{\textbf{Emotion Related}} \\
& Q44-Q45 & Views on technology development and respect for authority (1=Good, 2=Don't mind, 3=Bad) \\
& Q46 & Happiness level (1=Very happy to 4=Not at all happy) \\
& Q47 & Health status (1=Very good to 5=Very poor) \\
& Q48 & Freedom of choice and control over life (1=No choice to 10=Great deal of choice) \\
& Q49-Q50 & Life satisfaction and financial satisfaction (1=Completely dissatisfied to 10=Completely satisfied) \\
& Q52 & Felt unsafe from crime in last 12 months (1=Often to 4=Never) \\
& Q131 & General security feeling (1=Very secure to 4=Not at all secure) \\
& Q146-Q148 & Worry about war, terrorist attack, civil war (1=Very much to 4=Not at all) \\
\midrule
\multicolumn{3}{l}{\textbf{Heuristic}} \\
& Q94-Q104 & Active membership in: church/religious, sports/recreation, arts/music/education, labor union, political party, environmental, professional, humanitarian/charity, consumer, self-help/mutual aid, women's groups (1=Inactive, 2=Active, 3=Don't belong) \\
& Q201-Q208 & Media/communication usage: newspaper, TV news, radio news, mobile phone, email, internet, social media, talking with friends (1=Daily to 5=Never) \\
& Q18-Q26 & Unwanted neighbors: drug addicts, different race, AIDS, immigrants, homosexuals, different religion, heavy drinkers, unmarried couples, different language (1=Don't want, 2=Want) \\
& Q29-Q31 & Gender attitudes: men as better political leaders, university more important for boys, men as better business executives (1=Strongly agree to 4=Strongly disagree) \\
& Q33-Q35 & Job/gender attitudes: men's right to jobs when scarce, natives over immigrants, woman earning more causes problems (1=Strongly agree to 5=Strongly disagree) \\
\bottomrule
\end{tabular}
\end{table*}

\section{Significance Tests Across Experiments}
\noindent \textbf{Prompt-based Conditioning (susceptibility alignment)}: Conditioning LLMs with belief information improves demographic susceptibility simulation compared to zero-shot and demo-only prompts. We averaged our findings across 3 individual runs, and improvements of belief settings are statistically significant (paired t-test, p < 0.05 for both PANDORA and MIST-1). 

\noindent \textbf{Belief Dimension Analysis.} Certain belief dimensions are more predictive of susceptibility than others. We use repeated-measures ANOVA and performance variation vary significantly by belief dimensions, more for PANDORA, with a few non-significant variation for MIST (congitive, and alldimensions). 

\noindent \textbf{Post-Training Adaptation} In PANDORA+MIST-1, accuracy differences between baseline and the 2‑phase methods are modest (e.g., 0.75 vs. 0.793 for Llama), and z‑tests show that both BeliefSim-FT and LoRA fine‑tuning significantly outperforms the baseline (p=0.02–0.03). On MIST‑2 the gains are pronounced: for Llama the baseline accuracy is 0.680 while BeliefSim-FT reaches 0.884 and LoRA‑FT 0.876. Two‑proportion z‑tests show that both BeliefSim-FT and LoRA‑FT significantly outperform the baseline (p<0.001).

\section{Model Choices, Implementation Details and Computational Resources}

We evaluate a diverse set of instruction-tuned models to cover variation in model family, scale, and alignment behavior. Specifically, we include strong open-source chat models from different families (\texttt{Qwen}, \texttt{Llama}, \texttt{Mistral}, \texttt{OLMo}, \texttt{Grok})  and \texttt{DeepSeek}) to test whether belief-based conditioning is robust beyond a single architecture or training recipe. Due to resource constraints, we do not include substantially larger open-source models; instead, we focus on models in the 3B--14B range that support repeated prompting, ablations, and counterfactual evaluations at scale.  Finally, \texttt{Grok-4-Fast} provides a stronger closed-source comparison point. This model set therefore lets us compare both performance and demographic sensitivity across heterogeneous instruction-following systems.

We conduct experiments using instruction-tuned LLMs including Llama, Qwen, and Mistral, implemented with the Hugging Face Transformers and PEFT libraries. Our framework adopts a modular design with explicit belief heads and susceptibility heads, and employs LoRA-based fine-tuning where base-model adaptation is required, while relying on lightweight head training in BeliefSim-FT to decouple belief modeling from susceptibility prediction. All experiments are run on NVIDIA A40 GPUs with a maximum sequence length of 1024 tokens and fixed random seeds to ensure reproducibility and computational efficiency.

\section{Reproducibility}
We open-source our codes and data, which are uploaded to the submission system. This would help future work to reproduce our results and explore BeliefSim: demographic-aware misinformation susceptibility simulation in LLMs using belief-priors, and BeliefSim-FT.

\end{document}